\RequirePackage{fix-cm}
\documentclass[twocolumn]{svjour3}          
\smartqed  
\usepackage{graphicx}
\usepackage[utf8]{inputenc}
\usepackage[english]{babel}
\usepackage[sort]{natbib}
\usepackage{amsmath}
\usepackage{amssymb}
\usepackage{glossaries}
\usepackage{xspace}
\usepackage{xcolor}
\usepackage{enumitem}
\usepackage{multirow}
\usepackage{adjustbox}
\usepackage{hhline}
\usepackage{subcaption}

\makeatletter
\if@twocolumn
  \renewcommand\normalsize{%
   \@setfontsize\normalsize\@xpt{12.5pt}%
   \abovedisplayskip=3 mm plus6pt minus 4pt
   \belowdisplayskip=3 mm plus6pt minus 4pt
   \abovedisplayshortskip=0.0 mm plus6pt
   \belowdisplayshortskip=2 mm plus4pt minus 4pt
   \let\@listi\@listI}%

  \renewcommand\small{%
   \@setfontsize\small{8.5pt}\@xpt
   \abovedisplayskip 8.5\p@ \@plus3\p@ \@minus4\p@
   \abovedisplayshortskip \z@ \@plus2\p@
   \belowdisplayshortskip 4\p@ \@plus2\p@ \@minus2\p@
   \def\@listi{\leftmargin\leftmargini
               \parsep 0\p@ \@plus1\p@ \@minus\p@
               \topsep 4\p@ \@plus2\p@ \@minus4\p@
               \itemsep0\p@}%
   \belowdisplayskip \abovedisplayskip}

\else
  \if@smallext
   \renewcommand\normalsize{%
   \@setfontsize\normalsize\@xpt\@xiipt
   \abovedisplayskip=3 mm plus6pt minus 4pt
   \belowdisplayskip=3 mm plus6pt minus 4pt
   \abovedisplayshortskip=0.0 mm plus6pt
   \belowdisplayshortskip=2 mm plus4pt minus 4pt
   \let\@listi\@listI}%

  \renewcommand\small{%
   \@setfontsize\small\@viiipt{9.5pt}%
   \abovedisplayskip 8.5\p@ \@plus3\p@ \@minus4\p@
   \abovedisplayshortskip \z@ \@plus2\p@
   \belowdisplayshortskip 4\p@ \@plus2\p@ \@minus2\p@
   \def\@listi{\leftmargin\leftmargini
               \parsep 0\p@ \@plus1\p@ \@minus\p@
               \topsep 4\p@ \@plus2\p@ \@minus4\p@
               \itemsep0\p@}%
   \belowdisplayskip \abovedisplayskip}
 \else
  \renewcommand\normalsize{%
   \@setfontsize\normalsize{9.5pt}{11.5pt}%
   \abovedisplayskip=3 mm plus6pt minus 4pt
   \belowdisplayskip=3 mm plus6pt minus 4pt
   \abovedisplayshortskip=0.0 mm plus6pt
   \belowdisplayshortskip=2 mm plus4pt minus 4pt
   \let\@listi\@listI}%

  \renewcommand\small{%
   \@setfontsize\small\@viiipt{9.25pt}%
   \abovedisplayskip 8.5\p@ \@plus3\p@ \@minus4\p@
   \abovedisplayshortskip \z@ \@plus2\p@
   \belowdisplayshortskip 4\p@ \@plus2\p@ \@minus2\p@
   \def\@listi{\leftmargin\leftmargini
               \parsep 0\p@ \@plus1\p@ \@minus\p@
               \topsep 4\p@ \@plus2\p@ \@minus4\p@
               \itemsep0\p@}%
   \belowdisplayskip \abovedisplayskip}
  \fi
\fi

\makeatother
\usepackage[final]{microtype}

\usepackage[colorlinks,allcolors=blue]{hyperref}

\makeatletter \let\cl@chapter\relax \makeatother
\usepackage[sort&compress,capitalize,noabbrev,nameinlink]{cleveref}
\creflabelformat{equation}{#2#1#3}

\newcommand*{\eg}{e.g.\@\xspace}
\newcommand*{\ie}{i.e.\@\xspace}

\newcommand\Tstrut{\rule{0pt}{2.6ex}}         
\newcommand\Bstrut{\rule[-0.8ex]{0pt}{0pt}}   

\newacronym{wht}{WHT}{Walsh-Hadamard-Transform}
\newacronym{sgm}{SGM}{Semi-Global-Matching}
\newacronym{prsm}{PRSM}{Piece-wise Rigid Scene Model}
\newacronym{sed}{SED}{Structured Edge Detection}
\newacronym{sor}{SOR}{Successive Over-Relaxation}
\newacronym{sff}{SFF}{SceneFlowFields}
\newacronym{sffpp}{SFF++}{SceneFlowFields++}

\DeclareMathOperator*{\argmin}{arg\,min}

\begin{document}

\title{SceneFlowFields++:\\Multi-frame Matching, Visibility Prediction, and Robust Interpolation for Scene Flow Estimation
}

\titlerunning{SceneFlowFields++}        

\author{René Schuster         \and
		Oliver Wasenmüller \and
		Christian Unger \and
		Georg Kuschk \and
		Didier Stricker 
}


\institute{R. Schuster \and O. Wasenmüller \and D. Stricker \at
              DFKI - German Research Center for Artificial Intelligence\\
              Kaiserslautern, Germany\\
              \email{firstname.lastname@dfki.de}           
           \and
           C. Unger \and G. Kuschk \at
              BMW Group\\
              Munich, Germany\\
              \email{christian.unger@bmw.de}\\
              \email{g.kuschk@gmx.de}
}

\date{Received: date / Accepted: date}

\maketitle

\begin{abstract}
State-of-the-art scene flow algorithms pursue the conflicting targets of accuracy, run time, and robustness. With the successful concept of pixel-wise matching and sparse-to-dense interpolation, we shift the operating point in this field of conflicts towards universality and speed. Avoiding strong assumptions on the domain or the problem yields a more robust algorithm. This algorithm is fast because we avoid explicit regularization during matching, which allows an efficient computation. Using image information from multiple time steps and explicit visibility prediction based on previous results, we achieve competitive performances on different data sets. Our contributions and results are evaluated in comparative experiments. Overall, we present an accurate scene flow algorithm that is faster and more generic than any individual benchmark leader.

\end{abstract}

\begin{figure}[t]
	\centering
	\begin{subfigure}[c]{0.9\columnwidth}
		\includegraphics[width=1\textwidth]{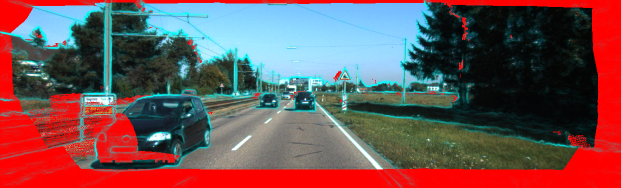}
		\caption{Reference image with predicted invisibility (red).}
		\label{fig:title:image}
		\vspace{1.6mm}%
	\end{subfigure}
	\begin{subfigure}[c]{0.9\columnwidth}
		\centering
		\begin{subfigure}[c]{1\columnwidth}
			\includegraphics[width=1\textwidth]{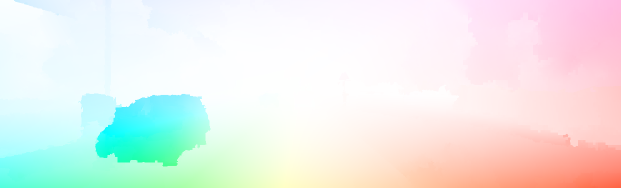}%
			\vspace{1mm}%
		\end{subfigure}
		\begin{subfigure}[c]{1\columnwidth}
			\includegraphics[width=1\textwidth]{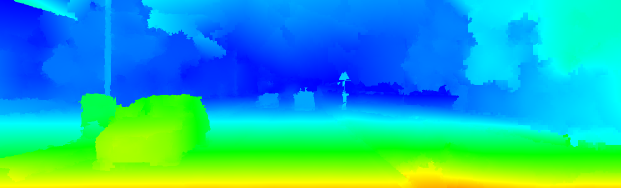}%
			\vspace{1mm}%
		\end{subfigure}
		\begin{subfigure}[c]{1\columnwidth}
			\includegraphics[width=1\textwidth]{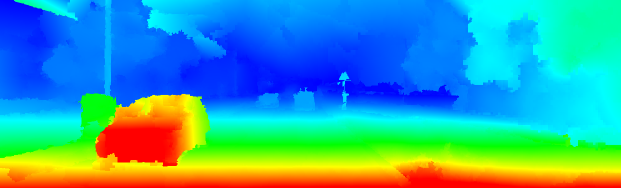}%
		\end{subfigure}
		\caption{Final result of SceneFlowFields++.}
		\label{fig:title:matching}
	\end{subfigure}		
	\caption{SceneFlowFields++ uses explicit visibility reasoning and the information of multiple frames to correctly estimate scene flow in occluded and out-of-bounds regions. Regularization is imposed by a consistency check and robust interpolation.}
	\label{fig:title}
\end{figure}

\section{Introduction} \label{sec:intro}
Scene flow is the problem of estimating the perceived dense 3D motion field along with the 3D geometry of the scene. Typically, a set of stereo image pairs is used to estimate scene flow.
Motion estimation is a crucial part for autonomous systems like self-driving vehicles and in many applications such as collision detection and path planning in driver assistance systems and robot navigation, object tracking, moving object detection, frame interpolation (temporal super resolution), and many others. 

Traditionally, motion is described by optical flow in 2D. Recently, with more computational resources and stereo cameras in mobile systems and vehicles available, the more complex, yet more realistic representation of motion in 3D is gaining increasing interest from research and industry. Despite growing efforts, either problem under unconstrained settings is far from being solved. However, with certain assumptions and restrictions to special domains, there exist methods that achieve impressive results on popular benchmarks.
Yet, any assumption -- even if it's reasonable for a certain domain -- will limit the applicability to other domains. The variance of performance of an algorithm on different data sets is one indicator for the robustness of that algorithm. Especially when applying deep neural networks, the generalization to different domains suffers heavily, unless the specific network is retrained or fine-tuned for the new domain. On the other hand, robustness alone should not be the only goal. A reasonable level of accuracy is required for any application that uses scene flow. Often, high accuracy comes at the price of run time. Top performing methods on the KITTI benchmark \citep{menze2015object} have run times of more than 5 minutes. Existing algorithms are focused on one or at most two of these targets, either speed, accuracy, or robustness. Upcoming deep learning approaches are potentially fast and accurate, but lack generalization. The state-of-the-art can be very accurate -- sometimes even across different domains -- but has high run times often.

With our novel concept of \gls*{sff} \citep{schuster2018sceneflowfields}, we want to push the limits of scene flow estimation towards a robust, comparatively fast, and accurate algorithm. Towards this end, we restrict ourselves to very few reasonable assumptions with minor impact on the versatility. Further, we avoid explicit regularization in our scene flow problem formulation which enables a fast computation of the solution. Regularization is imposed by outlier filtering and interpolation as it is often done in state-of-the-art optical flow algorithms \citep{schuster2018ffpp,hu2017robust,revaud2015epic,bailer2015flow}. This approach is fast and domain invariant. However, the basis is accurate matching based on visual similarity in images. Therefore, we have to deal with the typical problems of matching tasks: Changes in lighting, image noise, perspective deformations, occlusions, etc. 
To further improve the accuracy and robustness of this sparse-to-dense pipeline, we extend the concept to a multi-frame setup with explicit visibility reasoning. We also use novel concepts for interpolation to increase the robustness.

\begin{figure*}[t]
	\centering
	\includegraphics[width=0.95\linewidth]{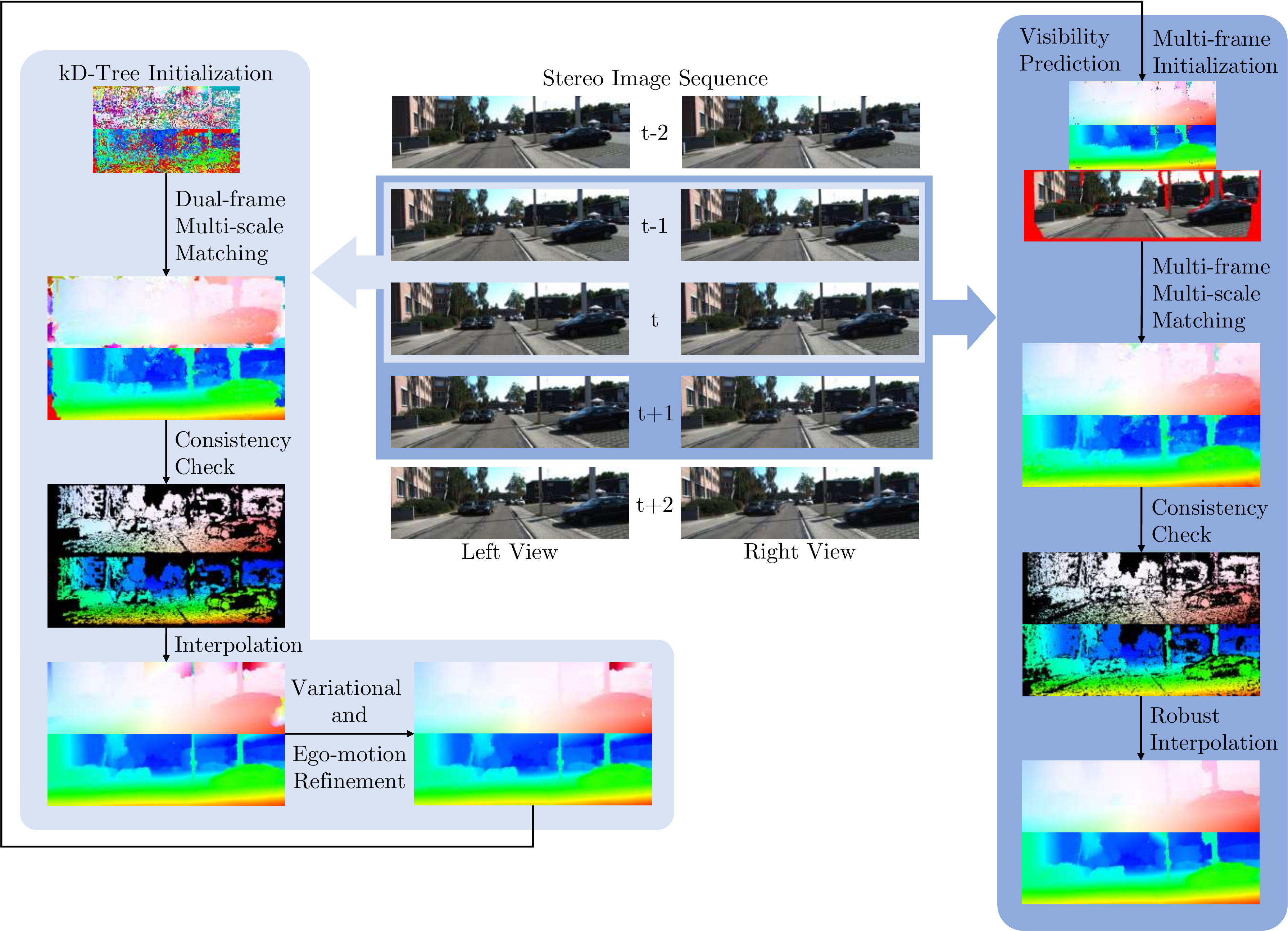}
	\caption{Overview of our dual- and multi-frame approach. We work on a sequence of stereo image pairs. \gls*{sff} (left) uses two frame pairs, initializes on a sub-scale and applies multi-scale propagation and random search for matching. The matched result is filtered in a consistency check, interpolated and refined by variational optimization and the optional ego-motion model. The final result is used to initialize the multi-frame process \gls*{sffpp} (right) which uses three frame pairs. Matching is done with explicit visibility reasoning over all images. The consistency check is adjusted to the novel setup, and the interpolation uses novel concepts for increased robustness. Note that there is no refinement necessary in our improved robust multi-frame pipeline. Each step is illustrated by the corresponding optical flow and one disparity map of the respective scene flow field. \Cref{tab:contrast} summarizes the differences between the dual-frame and multi-frame approach.}
	\label{fig:overview}
\end{figure*}

Working in the context of classical, image-based scene flow estimation, we assume to have consecutive frame pairs of a stereo image sequence (at least 4 images, cf. \cref{fig:overview}) as input. We represent scene flow as a 4D vector $\textbf{s} =(u,v,d_0,d_1)^T$ consisting of two optical flow components $u$, $v$ and the disparity values $d_0$, $d_1$ for both time steps. This way, we can express pixel correspondences in the four relevant images. Having rectified stereo image pairs and calibrated cameras, the 4D representation in image space is equivalent to the full scene flow representation of 3D geometry and 3D motion in world space. 

This paper re-presents our work of \gls*{sff} \citep{schuster2018sceneflowfields} and extends it considerably. In detail, the contributions of the original conference paper are:
\begin{itemize}[noitemsep,topsep=1pt,label={\tiny\raisebox{0.75ex}{\textbullet}}]
\item The first method of sparse-to-dense interpolation for scene flow.
\item A novel method to find scene flow matches.
\item A new interpolation method for scene flow that preserves boundaries of geometry and motion.
\item An improved edge detector to approximate scene flow boundaries for the KITTI data set.
\item An optional approach for straightforward integration of ego-motion.
\end{itemize}
We substantially extend \gls*{sff} in this paper by the following contributions:
\begin{itemize}[noitemsep,topsep=1pt,label={\tiny\raisebox{0.75ex}{\textbullet}}]
\item Novel matching and filtering for scene flow in a multi-frame instead of dual-frame setup.
\item Explicit one-shot visibility reasoning for scene flow matching on pixel-level.
\item More robust two-stage interpolation of scene flow correspondences.
\item A unified boundary detector for different data sets.
\item Thorough evaluation of our conference approach and our novel extension with comparison to state-of-the-art on KITTI and MPI Sintel.
\end{itemize}
Since the extensions are based on the original \gls*{sff}, we term our novel approach \gls*{sffpp}.

The rest of the paper is structured as follows: We give an overview of related work in \cref{sec:related}. \Cref{sec:sff} presents details about \gls*{sff}, followed by an incremental description of \gls*{sffpp} in \cref{sec:extension}. Experimental comparisons between \gls{sff} and \gls*{sffpp} and to state-of-the-art are described in \cref{sec:results}. We summarize our findings in \cref{sec:conclusion}.

\section{Related Work} \label{sec:related}
\paragraph{Data Sets.}
Since it is hard to capture ground truth scene flow information, there exist only very few data sets to evaluate scene flow algorithms on. Most of them use virtually rendered scenes to obtain the ground truth data \citep{butler2012sintel,gaidon2016vkitti,mayer2016large}. To the best of our knowledge the only realistic data set that provides a benchmark for scene flow is the KITTI Vision Benchmark \citep{geiger2012kitti} that combines various tasks for automotive vision. Its introduction has played an important role in the development of stereo and optical flow algorithms, and the extension by \citet{menze2015object} has also driven the progress in scene flow estimation.

\paragraph{Variational Scene Flow Estimation.}
\Citet{vedula1999three} were among the first to compute 3D scene flow. Afterwards, many variational approaches for scene flow estimation followed. First using pure color images as input \citep{basha2013multi,huguet2007variational} and later using RGB-D images \citep{herbst2013rgbd,jaimez2015primal,wedel2008efficient}. While a variational formulation is typically complex, \citet{jaimez2015primal} achieved real-time performance with a primal-dual framework. Yet, all these approaches are sensitive to initialization and can not cope with large displacements, which is why they use a coarse-to-fine scheme. That in turn tends to miss finer details. Furthermore, the RGB-D approaches rely on depth sensors that either perform poorly in outdoor scenarios or are accordingly very expensive \citep{yoshida2017time}. Nowadays, variational methods are outperformed in terms of speed and accuracy by other approaches and are only used as a refinement step.

\paragraph{Combination Approach.}
Because the scene flow problem is highly related to the auxiliary tasks of optical flow and stereo disparity estimation, people have tried to estimate scene flow by combining separate results for optical flow and stereo disparity \citep{schuster2018combining}. Though the separation brings advantages for the complexity of the problem and thus the run time, it is believed that a single formulation of the problem yields more consistent scene flow results, which are more accurate and realistic.

\paragraph{Assumption of Planarity and Rigidity.}
Due to the advent of a piece-wise rigid plane model \citep{vogel2013PRSF}, scene flow has recently achieved a boost in performance. The majority of top performing methods at the KITTI Vision Benchmark employ this model to enforce strong regularization \citep{neoral2017object,behl2017bounding,lv2016CSF,menze2015object,vogel2015PRSM}. 
\Citet{vogel2014view,vogel2015PRSM} encode this model by alternating assignment of each pixel to a plane segment and each segment to a rigid motion, based on a discrete set of planes and motions in view-consistent manner over multiple frames. \Citet{neoral2017object,behl2017bounding,menze2015object} further lower the complexity of the model by the assumption that a scene consists of very few independently moving rigid objects. Thus each plane segment only needs to be assigned to one object. All segments assigned to the same object share the same motion. By propagation of objects over multiple frames, \citet{neoral2017object} achieve temporal consistency for the model of \citet{menze2015object}. \Citet{behl2017bounding} use deep learning to obtain semantic object information a-priori. \Citet{lv2016CSF} solve the pixel-to-plane assignment and the plane-to-motion assignment in a continuous domain.
Despite the remarkable accuracy on KITTI, many of these methods are not applicable to domains with different characteristics. The rigid motion assumption is strongly violated by articulated gestures and other non-rigid motions that often occur in the Sintel data set \citep{butler2012sintel}. The assumption made by \citet{menze2015object,neoral2017object,behl2017bounding} that there are only a few independent dynamic objects in a scene is inappropriate for highly dynamic scenarios. Further, methods falling into this category typically have very long run times of several minutes up to almost one hour per frame.

\paragraph{Guided by Semantics.}
Other scene flow algorithms use deep learning to incorporate semantic information into the motion estimation problem \citep{ren2017cascaded}. Yet, in terms of robustness, deep learning approaches \citep{behl2017bounding,ren2017cascaded} typically lag behind because they generalize badly to unseen data and even worse to data from different domains \citep{wang2018deep}. This is especially true for semantic segmentation, where the domain gap is amplified by the mismatch of semantic classes between domains \citep{lv2018domain}.
Our algorithm is especially designed to achieve good results across different data sets without any dedicated parameter tuning.

\paragraph{Static-Dynamic Decomposition.}
Yet another promising strategy builds on the decomposition of a scene into static and moving parts \citep{taniai2017fsf}. While the motion of dynamic objects is estimated by solving a discrete labeling problem (as by \citet{chen2016full}) using the \gls*{sgm} \citep{hirschmuller2008SGM} algorithm, the perceived motion of all static parts is directly obtained from the 3D geometry of the scene and the ego-motion of the camera. This approach is especially convenient for scenes, where only a small proportion consists of moving objects, like it is usually the case in traffic scenarios.
However, any a-priori assumption limits the versatility of a method. A rigid plane model performs poorly when applied to deformable objects, and ego-motion estimation for highly dynamic scenes is hard.

\begin{table*}[t]
	\centering
	\caption{Comparison of different categories of scene flow algorithms.}
	\label{tab:categories}
	\resizebox{0.9\textwidth}{!}{\begin{tabular}{c|cccc}
		Category &  Advantages & Disadvantages & Examples\Bstrut\\
		\hhline{=|====}
		Variational &  & \begin{tabular}{c} Slow, inaccurate,\\ only indoors\end{tabular} & \begin{tabular}{c} \citep{huguet2007variational}\Tstrut\\ \citep{herbst2013rgbd}\Bstrut\end{tabular}\Tstrut\Bstrut\\
		\hline
		Decomposition & \begin{tabular}{c} Potentially\\fast \end{tabular} & \begin{tabular}{c} Ego-motion\Tstrut\\ dependency,\\ inconsistent\Bstrut\end{tabular} & \begin{tabular}{c} \citep{schuster2018combining}\\ FSF \citep{taniai2017fsf}\end{tabular}\Tstrut\Bstrut\\
		\hline
		\begin{tabular}{c} Pice-wise Rigid\\ Planes Model\end{tabular} & \begin{tabular}{c} Strong\\ regularization\end{tabular} & Slow & \begin{tabular}{c} PRSF \citep{vogel2013PRSF}\Tstrut\\ OSF \citep{menze2015object}\\ CSF \citep{lv2016CSF}\Bstrut\end{tabular}\Tstrut\Bstrut\\
		\hline
		Deep Learning & Fast & \begin{tabular}{c} Poor\\ generalization\end{tabular} & \begin{tabular}{c} PWOC-3D \citep{saxena2019pwoc}\Tstrut\\ DRISF \citep{ma2019drisf}\Bstrut\end{tabular}\Tstrut\Bstrut\\
		\hline
		Sparse-to-Dense &  \begin{tabular}{c} Comparatively\Tstrut\\ fast, good\\generalization\Bstrut\end{tabular} & \begin{tabular}{c} Sensitive to\Tstrut\\distribution\\of matches\end{tabular}  & \begin{tabular}{c} SFF \cite{schuster2018sceneflowfields}\\ SFF++ (ours)\end{tabular}\Tstrut\\
	\end{tabular}}
\end{table*}

\paragraph{From Dual to Multiple Frames.}
When speaking about the related work, one has to differentiate between dual-frame \citep{vogel2013PRSF,lv2016CSF,menze2015object} and multi-frame \citep{neoral2017object,taniai2017fsf,vogel2015PRSM} approaches. Especially in the context of traffic scenarios, we have several characteristics that make matching between two frame pairs much more challenging than in a multi-view setting. These characteristics are: 1.) Considerably large stereo and flow displacements. 2.) Difficult lighting conditions and many reflective and (semi-)transparent surfaces of cars. 3.) Fast ego-motions sometimes combined with low frame rates, which causes large regions to move out of image bounds. 
This explains why there exist pairs of dual-frame and multi-frame methods. The transitions from OSF \citep{menze2015object} to OSF+TC \citep{neoral2017object} and from PRSF \citep{vogel2013PRSF} to \gls*{prsm} \citep{vogel2014view,vogel2015PRSM} brought essential improvements by using the additional information of multiple image frames.
However, this additional information comes at a cost. The relationship between multiple temporal steps needs to be modeled to make use of the additional images. A typical model is to assume smooth, constant motion between neighboring time steps \citep{neoral2017object,vogel2015PRSM}.
In this extension of \gls*{sff} \citep{schuster2018sceneflowfields}, we also follow this trend and use more than two image pairs, assuming constant motion.

\paragraph{Discrimination from State-of-the-Art.}
We can cluster related work into certain categories (cf. \cref{tab:categories}).
Because our scene flow method follows the newly introduced sparse-to-dense approach, it differs from any of the related approaches. We find sparse scene flow matches that are interpolated to a dense scene flow field, recovering the geometry of the scene and the 3D motion. 
Our method has to be distinguished from purely variational approaches. Although we use variational optimization, it can be considered as a post-processing step for refinement.
Similar, the cameras ego-motion is only used to refine the results for static image regions.
During interpolation, we assume that the geometry of a scene can be modeled by very small planar segments, but we do not initially presume any coarse segmentation. In fact, the very small size of our plane segments leads to smoothly curved shapes and sharp boundaries. The same holds for our piece-wise motion model that is used to interpolate the 3D motion.
Methods which are guided by semantic segmentation from deep neural networks will generalize badly to other domains, unless they are fine-tuned for the new task. Same is assumed for upcoming purely learning based approaches \citep{ma2019drisf,saxena2019pwoc} which are potentially even faster than our approach.  \gls*{sffpp} focuses especially on robustness across domains and applications.

We contrast the different categories and compare their properties in \cref{tab:categories}.
The piece-wise rigid planes models is particularly accurate due to its strong regularization (as long as the assumptions are not violated), but is also complex and computationally expensive.
The decomposition (separation) approach that splits the scene flow problem into less difficult sub-problems is especially fast and benefits from advances in the auxiliary tasks. Yet, separate computation leads to overall less consistent scene flow.
Deep learning is potentially fast due to the inherent parallelization on GPUs, but sensitive to the distribution of available training data and not interpretable in case of failure.
The novel scene flow concept of sparse-to-dense interpolation allows to separate matching from regularization. With the use of appropriate interpolation models and interpolation regions, the negative impact of violated assumptions can be diminished.
However, the separation of matching and regularization makes our approach sensitive to the quality of the sparse matching results. To overcome this issue, we propose to use multiple frames within this work.

\section{SceneFlowFields} \label{sec:sff}
For our dual-frame scene flow estimation we assume to have the typical stereo image information provided, \ie two rectified stereo image pairs ($I_0^l, I_1^l, I_0^r, I_1^r$) at times $t$ and $t+1$ along with the camera intrinsics. We further assume that the baseline $B$ is known. For rectified images, the baseline describes the relative pose between the left and right cameras as translation parallel to the image plane. 
During matching, we jointly optimize all four components to obtain coherent scene flow. Our matching algorithm follows the ideas of non-regularized, coarse-to-fine matching with propagation and random search \citep{bailer2015flow,bailer2018flowfields} which was shown to work the best for sparse-to-dense optical flow pipelines \citep{bailer2015flow,zweig2017interponet,schuster2018ffpp}.
Given the mentioned information, we estimate a dense scene flow field as follows: 
For $k$ sub-scales we initialize the coarsest scale. For all $k+1$ scales (the $k=3$ sub-scales plus full resolution), we iteratively propagate scene flow vectors and adjust them by random search. Afterwards, the dense scene flow map on full resolution is filtered using a two-stage consistency check. The filtered scene flow map is further thinned out in a sparsification step. Scene flow boundaries are detected using a structured random forest. Geometry and 3D motion are separately interpolated with respect to object boundaries of the scene. Finally, we refine the 3D motion by variational optimization. An overview of the dual-frame method \gls*{sff} is outlined in \cref{fig:overview}. 

\subsection{Sparse Correspondences} \label{sec:sff:matching}
\paragraph{Matching Cost.}
The matching cost in our algorithm solely depends on a data term. No additional smoothness assumptions are made like \eg by \citet{herbst2013rgbd,huguet2007variational,lv2016CSF,menze2015object,vogel2013PRSF,vogel2015PRSM}. Given a scene flow vector, we define its matching cost by the sum of Euclidean distances between SIFTFlow features \citep{liu2011siftflow} over small image patches.
Our matching error for two corresponding pixels $\mathbf{p}$ and $\mathbf{p'}$ in images $I$ and $I'$ is defined by the following cost
\begin{multline} \label{eq:matchingcost}
	C\left(I,\mathbf{p},I',\mathbf{p'}\right) = \\ 
	\sum_{\mathbf{\tilde{p}} \in W(\mathbf{p})} \left\lVert \phi\left(I,\mathbf{\tilde{p}}\right) - \phi\left(I',\mathbf{p'}+\mathbf{\tilde{p}}-\mathbf{p}\right) \right\rVert,
\end{multline}
where $W(\mathbf{p})$ is a $7 \times 7$ patch window centered at pixel $\mathbf{p}$ and $\phi\left(I,\mathbf{p}\right)$ is a function that returns the first three principal components of a SIFT feature vector (SIFTFlow) for pixel $\mathbf{p}$ in image $I$.
We evaluate the cost for three image correspondences. The stereo image pair at time $t$, the temporal image pair for the left view point (standard optical flow correspondence) and a cross correspondence between the reference frame and the right frame at the next time step (cf. \cref{fig:overview}). This leads to the following overall cost $C$ for a scene flow vector $\mathbf{s} = \left(u,v,d_0,d_1\right)^T$ at pixel $\mathbf{p}$:
\begin{equation} \label{eq:cost_dual}
\begin{split}
	C\left(\mathbf{p},\mathbf{s}\right) &= C\left(I_0^l,\mathbf{p},I_1^l,\mathbf{p} +\left(u,v\right)^T\right)\\
	& + C\left(I_0^l,\mathbf{p},I_0^r,\mathbf{p}+\left(-d_0,0\right)^T\right)\\
	& + C\left(I_0^l,\mathbf{p},I_1^r,\mathbf{p}+\left(u-d_1,v\right)^T\right).
\end{split}
\end{equation}
With that we can obtain a dense scene flow field $\mathcal{S}$ by optimizing the following energy minimization problem
\begin{equation} \label{eq:energy}
	E\left(\mathbf{I}, \mathcal{S}\right) = \int_\Omega C\left(\mathbf{p},\mathcal{S}\left(\mathbf{p}\right)\right)\text{d}\mathbf{p}.
\end{equation}
\begin{equation} \label{eq:minimization}
	\hat{\mathcal{S}} = \argmin_{\mathcal{S}}\ E\left(\mathbf{I},\mathcal{S}\right).
\end{equation}

Though this optimization includes a lot of variables, we can exploit the fact that we can optimize $\mathbf{s} = \left(u,v,d_0,d_1\right)^T$ for each pixel individually since our formulation includes no inter-pixel dependencies, e.g. no explicit regularization.
Therefore, we use an efficient, greedy, stochastic optimization approach of propagation and random search.

\paragraph{Initialization.} 
Initialization is based on kD-trees similar to the work of \citet{he2012computing}, but with three trees, using \gls*{wht} features as done by  \citet{bailer2015flow,wannenwetsch2017probflow,schuster2018ffpp}. For each frame other than the reference frame ($I_0^l$), we compute a \gls*{wht} feature vector per pixel and store them in a kD-tree. To initialize a pixel of the reference image, we query the feature vector of the pixel to the pre-computed kD-trees. Scene flow matches are then obtained by comparing all combinations of the leafs for each queried node according to the matching data term introduced before (\cref{eq:cost_dual}). Since our stereo image pairs are rectified, for the images observed from the right camera view, we create kD-trees which regard the epipolar constraint, \ie queries for such a tree will only return elements, which lie on the same image row as the query pixel. This way, we can efficiently lower the number of leaves per node for the epipolar trees, which speeds up the initialization process without loss of accuracy. For further acceleration, we use this initialization on the coarsest resolution only, and let the propagation fill the gaps when evolving to the next higher scale.

Though the result of the initialization looks very noisy and unreliable (cf. \cref{fig:overview}), it is much better than random initialization. In fact, we could not reproduce similar results of our full approach with random initialization or any other more simple initialization method. This is because although the \gls*{wht} features are not very expressive, they are sufficient to find at least a few close-to-correct initial scene flow values, which later are spread and refined.

\paragraph{Multi-Scale Propagation.} 
The initial matches are spread by propagation and steadily refined by random search. This is done over multiple scales which helps to distribute rare correct initial matches over the whole image. For each scale, we run several iterations of propagation in one out of the four image quadrants, so that each direction is used equally often. During propagation, a scene flow vector will be replaced if the propagated vector has a smaller matching cost (see \cref{eq:cost_dual}). If this is not the case, the propagation along this path will continue with the previously existing scene flow vector. After each iteration we perform a random search. This means that for all pixels we add a uniformly distributed random offset in the interval $\left]-1,1\right[$ in pixel units of the current scale to each of the four scene flow components and check whether the matching cost decreases. Both propagation and random search help to obtain a smoothly varying vector field and to find correct matches even if the initialization is flawed. For the different scale spaces, we simulate the scaling by smoothing the images and taking only every $n$-th pixel for a subsampling factor of $n = 2^k$, so that the patches consist of the same number of pixels for all scales. This way, we prevent (up)sampling errors because all operations are performed on exact pixel locations on the full image resolution. Smoothing is done by area-based downsampling followed by upsampling using Lanczos interpolation. Note that a similar matching strategy has already been used by \citet{bailer2015flow}, but while they use it for optical flow, we apply it to twice as many dimensions in search space.
This matching method was shown to work very well for motion fields. Even unique initial values are spread across the whole image yielding smooth motion fields with sharp boundaries.

\paragraph{Consistency Check.}
The matching procedure yields a dense map of scene flow correspondences across all images. However, many of the correspondences are wrong because of occlusions, out-of-bounds motion or simply because of mismatching due to challenging image conditions. To remove these outliers, we perform a two-step consistency check. First, we compute an inverse scene flow field for which the reference image is the right image at time $t+1$. Temporal order as well as points of view are swapped. Everything else remains as explained above.

During consistency check, optical flow and both disparity maps for each pixel are compared to the corresponding values of the inverse scene flow field. If either difference exceeds a consistency threshold $\tau_c=1$ in image space (pixels), the complete scene flow vector will be removed. In a second stage, we form small regions of the remaining pixels as proposed by \citet{bailer2015flow}, where a pixel is added to a region if it has approximately the same scene flow vector. Afterwards, we check whether it is possible to add one of the already removed outliers in the neighborhood following the same rule. If this is possible and the region is smaller than $s_c=100$ pixels, we will remove the whole region. This way we obtain the filtered final scene flow correspondences of high accuracy and very few outliers (cf. \cref{tab:dual_vs_multi}). 

\begin{figure}[t]
	\begin{center}
		\begin{subfigure}[c]{0.475\columnwidth}
			\includegraphics[width=1\textwidth]{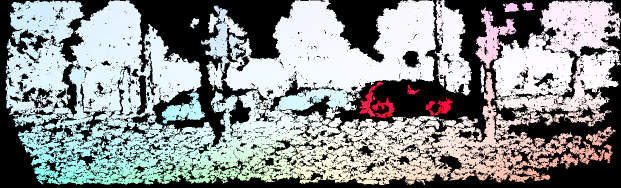}%
			\vspace{0.5mm}%
		\end{subfigure}
		\begin{subfigure}[c]{0.475\columnwidth}
			\includegraphics[width=1\textwidth]{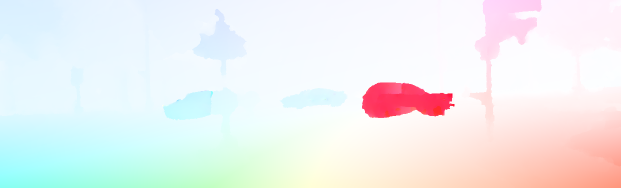}%
			\raisebox{2pt}{\makebox[0pt][r]{\bf a)~}}%
			\vspace{0.5mm}%
		\end{subfigure}
		\begin{subfigure}[c]{0.475\columnwidth}
			\includegraphics[width=1\textwidth]{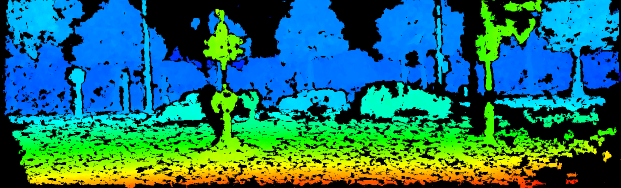}%
			\vspace{0.5mm}%
		\end{subfigure}
		\begin{subfigure}[c]{0.475\columnwidth}
			\includegraphics[width=1\textwidth]{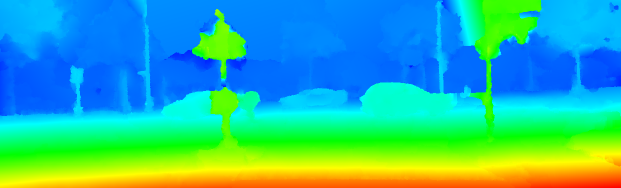}%
			\raisebox{2pt}{\makebox[0pt][r]{\bf b)~}}%
			\vspace{0.5mm}%
		\end{subfigure}
		\begin{subfigure}[c]{0.475\columnwidth}
			\includegraphics[width=1\textwidth]{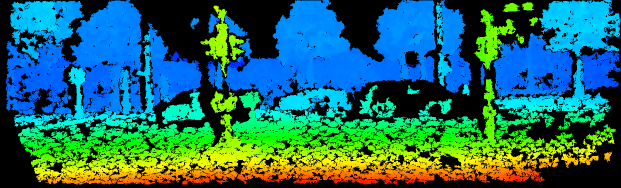}
		\end{subfigure}
		\begin{subfigure}[c]{0.475\columnwidth}
			\includegraphics[width=1\textwidth]{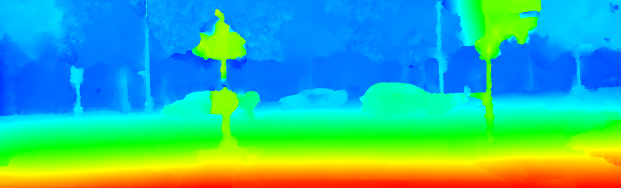}%
			\raisebox{2pt}{\makebox[0pt][r]{\bf c)~}}%
		\end{subfigure}
	\end{center}
	\caption{Sparse correspondences (left) and dense interpolation (right). Optical flow {\bf (a)} and disparities at $t$ {\bf (b)} and $t+1$ {\bf (c)}.}
	\label{fig:sparse-to-dense}
\end{figure}

Since we use a single scene flow formulation during consistency check also, the filtered results are very coherent. However, the joint filtering of all four scene flow components removes entire matches which might be only partly inconsistent. This applies especially to the reference disparity $d_0$. Since our interpolation model (see \cref{sec:sff:interpolation}) is split into separate steps for geometry and motion, we counter this problem, and fill up some gaps of the filtered disparity with additional values. These values are the result of a second independent consistency check for the disparity matches only. For the separate check we compute a second disparity map with \gls*{sgm} \citep{hirschmuller2008SGM} and use the same threshold $\tau_c$ as before. The additional disparity values that are retrieved this way are as accurate as the one from our standard consistency check, but much denser, which is shown in \cref{fig:sparse-to-dense}. We merge the result of the second consistency check for disparity only with the result of the first full consistency check.

\subsection{Dense Interpolation} \label{sec:sff:interpolation}
\paragraph{Sparsification.}
Before interpolating the filtered scene flow field to recover full density, an additional sparsification step is performed. This helps to extend the spatial support of the neighborhood during interpolation and speeds up the whole process \citep{bailer2015flow}. It also serves as a final filtering step, because for each non-overlapping $3 \times 3$ block, we select the match with the lowest consistency error during filtering only. The remaining matches are called seeds with respect to the interpolation.

\begin{figure}[t]
	\begin{center}
		\begin{subfigure}[c]{1\linewidth}
			\centering
			\resizebox{0.95\linewidth}{!}{
				\includegraphics[height=0.3\linewidth]{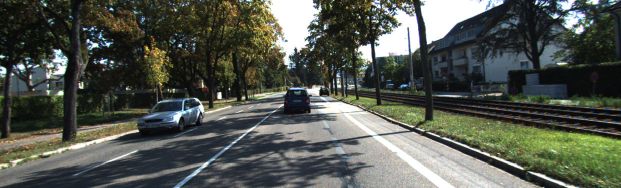}
				\includegraphics[height=0.3\linewidth]{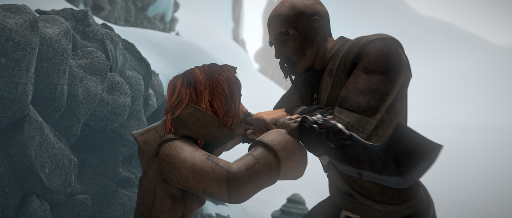}
			}
			\caption{Input images from different data sets.}
			\label{fig:edges:images}
			\vspace{1mm}%
		\end{subfigure}
		\begin{subfigure}[c]{1\linewidth}
			\centering
			\resizebox{0.95\linewidth}{!}{
				\includegraphics[height=0.3\linewidth]{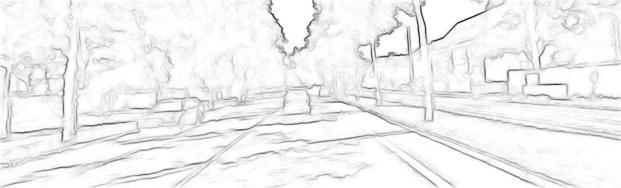}
				\includegraphics[height=0.3\linewidth]{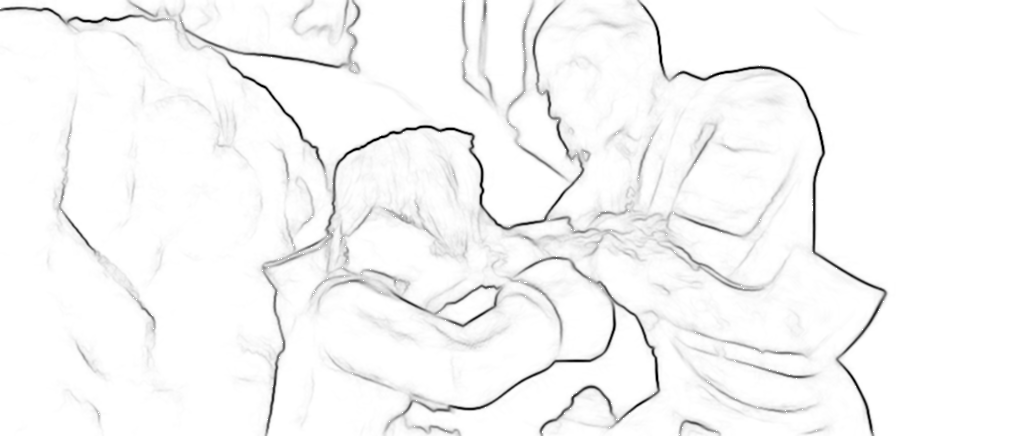}
			}
			\caption{Edge result from SED trained on BSDS.}
			\label{fig:edges:bsds}
			\vspace{1mm}%
		\end{subfigure}
		\begin{subfigure}[c]{1\linewidth}
			\centering
			\resizebox{0.95\linewidth}{!}{
				\includegraphics[height=0.3\linewidth]{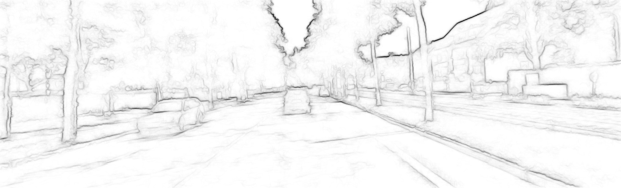}
				\includegraphics[height=0.3\linewidth]{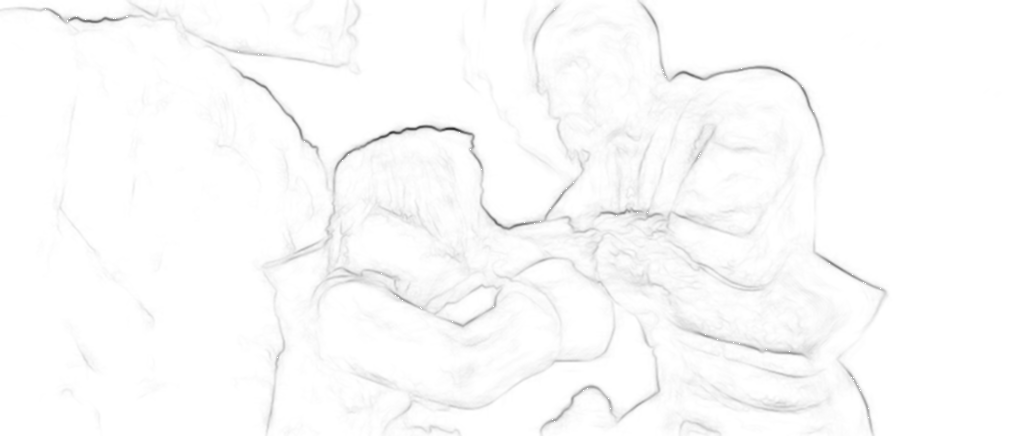}
			}
			\caption{Edges obtained using the KITTI model.}
			\label{fig:edges:kitti}
			\vspace{1mm}%
		\end{subfigure}
		\begin{subfigure}[c]{1\linewidth}
			\centering
			\resizebox{0.95\linewidth}{!}{
				\includegraphics[height=0.3\linewidth]{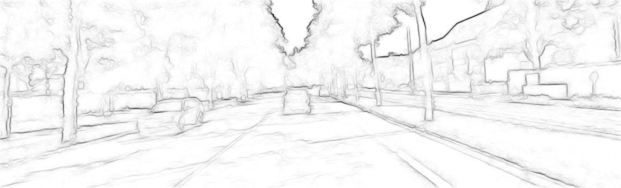}
				\includegraphics[height=0.3\linewidth]{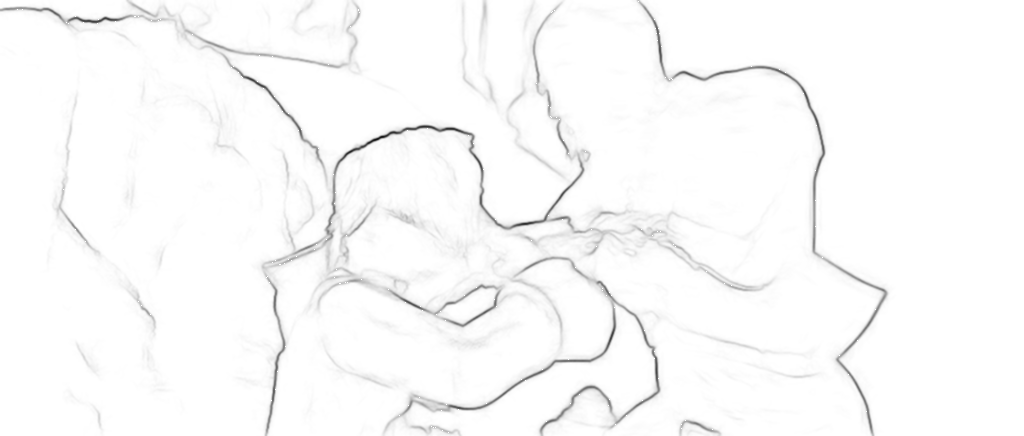}
			}
			\caption{Results from our unified boundary detector.}
			\label{fig:edges:mixed}
		\end{subfigure}
	\end{center}
\caption{Whereas \gls*{sed} \citep{dollar2013sed} (\subref{fig:edges:bsds}) detects all image gradients, our KITTI boundary detector (\subref{fig:edges:kitti}) suppresses lane markings and shadows. Our unified detector of \cref{sec:extension:interpolation} (\subref{fig:edges:mixed}), achieves a good trade-off between the advantages of (\subref{fig:edges:kitti}) and generalization abilities across different domains.}
\label{fig:edges}
\end{figure}

\paragraph{Interpolation Boundaries.}
A crucial part of the interpolation is the estimation of scene flow boundaries. While \citep{bailer2015flow,revaud2015epic} approximate motion boundaries for optical flow with a texture-agnostic edge detector \citep{dollar2013sed}, our edge detector is trained on semantic boundaries. We find that this models geometric boundaries, as well as motion boundaries, much better than image edges and is much more robust to lighting, shadows, and coarse textures. To do so, we have gathered 424 images of the KITTI data set from \citep{osep2016multi,ros2015offline,xu2016multimodal} that have been densely labeled with semantic class information. Within these images we have merged semantic classes that in general neither align with geometric nor motion discontinuities, \eg lane markings and road, or pole and panel.
The boundaries between the remaining semantic labels are used as binary edge maps to train our edge detector. To this end, we utilize the framework of \gls*{sed} \citep{dollar2013sed} and train a random forest with the same parameters as in their paper, except for the number of training patches. We sample twice as many positive and negative patches during training because we use a bigger data set with images of higher resolution. The impact of the novel boundary detector will be evaluated in our ablation study in \cref{sec:results:ablation} and is visualized in \cref{fig:edges}.

\paragraph{Interpolation Models.}
We interpolate geometry and 3D motion separately. This allows us to incorporate domain knowledge into the interpolation process by having dedicated models for both types of interpolation. Also, due to the separate consistency check for disparity only and full scene flow, we have different amounts of matches for geometry and motion, which favors the separation.
Apart from the models and number of seeds, the general concept of interpolation is the same. Suppose a local, boundary-aware neighborhood of 160 and 80 seeds is given for each unknown scene flow vector $\mathbf{\hat{s}}$ at pixel $\hat{p}$ for geometric and motion seeds respectively, $\mathcal{N}_{geo}$ and $\mathcal{N}_{motion}$. The depth of pixel $\hat{p}$ is reconstructed by fitting a plane $E(\hat{p}): a_1 x + a_2 y + a_3 = d_0$ through all seeds of the neighborhood $\mathcal{N}_{geo}$. This is done by solving a linear system of equations for all neighboring seed points $p_g$ for which the disparity values are known, using weighted least squares. The weights for each seed are obtained from a Gaussian kernel $g(D)= \exp{(-\alpha D)}$ on the distance $D(\hat{p},p_g)$ between target pixel and seed with coefficient $\alpha = 2.2$. The missing disparity value of $\hat{p}$ is obtained by plugging the coordinates of $\hat{p}$ into the estimated plane equation. In a similar fashion, but using a neighborhood of motion seeds $\mathcal{N}_{motion}$, the missing 3D motion is obtained by fitting an affine 3D transformation $\mathbf{x_1} = \mathbf{A} \mathbf{x_0} + \mathbf{t}$ using weighted least squares on all neighboring motion seeds $p_m$. Where $\mathbf{x_t} = (x_t,y_t,z_t)^T$ are the 3D world coordinates of motion seed $p_m$ at time $t$ and $t+1$, and $\left[\mathbf{A}|\mathbf{t}\right] \in \mathbb{R}^{3 \times 4}$ is the affine 3D transformation of twelve unknowns. The weights are computed by the same Gaussian kernel as for geometric interpolation, but using the distances $D(\hat{p},p_m)$ between the target pixel and the motion seeds. To summarize, for the full reconstruction of scene flow $\mathbf{\hat{s}}=(u,v,d_0,d_1)^T$ at pixel $\hat{p}$, we compute $d_0$ using the plane model $E(\hat{p})$, reproject the point into 3D world space, transform it according to its associated affine transformation $\left[\mathbf{A}|\mathbf{t}\right]$, and project it back to image space to obtain $u$, $v$ and $d_1$.

\paragraph{Edge-Aware Neighborhood.}
To find the local neighborhoods, we follow the idea of \citet{revaud2015epic} using both of their approximations. That is first, the $n$ closest seeds to a pixel $\hat{p}$ are the $n-1$ closest seeds to the closest seed of $\hat{p}$, thus all pixels with the same closest seed share the same local neighborhood. And secondly, the distance between $\hat{p}$ and its closest seed is a constant offset for all neighboring seeds, which can be neglected. It is therefore sufficient to find a labeling that assigns each pixel to its closest seed and to find the local neighborhood for each seed. We use the graph-based method of \citet{revaud2015epic} for this. The graph is constructed using the geodesic distances  between seeds, that are directly based on the edge maps from our boundary detector. A strong boundary, \ie a high value in the edge map, indicates a high cost for crossing that pixel, which leads to a high geodesic distance. The algorithm finds the shortest path between seeds and sums up the edge values along this path to obtain the geodesic distances between seeds.

\subsection{Variational Optimization} \label{sec:variational}
To further refine the 3D motion after interpolation, we use variational energy minimization to optimize the objective
\begin{equation} \label{eq:energy_var}
E(u,v,d') = E_{data}^{flow} + E_{data}^{cross} + \varphi \cdot E_{smooth}.
\end{equation}
Motion is represented in image space by optical flow and the change in disparity $d'$. The energy consists of three parts. Two data terms, one temporal correspondence and one cross correspondence, and an adaptively weighted smoothness term for regularization. The data terms use the gradient constancy assumption. Our experiments have shown, that a term for the color constancy assumption can be neglected.
\begin{multline} \label{eq:dataterm}
E_{data}^{\ast}(I,I',\mathbf{x},\mathbf{w}) = \\
\int_{\Omega} \beta\left(\mathbf{x},\mathbf{w}\right) \cdot \Psi\left( \gamma\cdot\left|\nabla I'(\mathbf{x}+\mathbf{w})-\nabla I(\mathbf{x})\right|^2\right)dx
\end{multline}
with $\mathbf{w}$ being a placeholder for either the optical flow displacement in \cref{eq:flowdata} or the cross image displacement in \cref{eq:crossdata}.
\begin{equation} \label{eq:flowdata}
E_{data}^{flow} = E_{data}^{\ast}\left(I_0^l,I_1^l,\mathbf{x},(u,v)^T\right)
\end{equation}
\begin{equation} \label{eq:crossdata}
E_{data}^{cross} = E_{data}^{\ast}\left(I_0^l,I_1^r,\mathbf{x},(u-d_0-d',v)^T\right)
\end{equation}
The data terms do not contribute to the energy if the function
\begin{equation}
\beta\left(\mathbf{x},\mathbf{w}\right) = \begin{cases}
1, &\text{if} \left(\mathbf{x}+\mathbf{w}\right)^T \in \Omega \\
0, &\text{otherwise}
\end{cases}
\end{equation}
indicates that the scene flow is leaving the image domain (out-of-bounds motion). The smoothness term 
\begin{equation} \label{eq:smoothness}
E_{smooth} = \int_{\Omega} \Psi\left(\left|\nabla u\right|^2 + \left|\nabla v\right|^2 + \lambda \cdot \left|\nabla d'\right|^2\right)dx
\end{equation}
penalizes changes in the motion field where $\left|\nabla u\right|^2$ is the magnitude of the spatial derivatives of one scene flow component. It is weighted by 
\begin{equation} \label{eq:smoothnessweight}
\varphi(\mathbf{x}) = e^{-\kappa B(\mathbf{x})},
\end{equation}
where $B(\mathbf{x})$ is the edge value of our boundary detector at pixel $\mathbf{x}$ (cf. \cref{sec:sff:interpolation}) and $\kappa=5$ is a kernel coefficient. All parts use the Charbonnier penalty $\Psi\left(x^2\right) = \sqrt{x^2+\varepsilon^2}$ with $\varepsilon^2=0.0001$ to achieve robustness. Since the smoothness term rather enforces constancy instead of smoothness if $\beta$ for both data terms is zero, we do not optimize the scene flow at pixels where the interpolated scene flow field leaves $\Omega$. Our energy formulation is inspired by \citep{brox2004high,huguet2007variational,wedel2008efficient}.
We use linear approximations of the Euler-Lagrange equations for the objective in \cref{eq:energy_var} and apply the framework of \citet{brox2004high} without the coarse-to-fine steps to find a solution by \gls*{sor}. Again, we will compare the results with and without the variational refinement in our ablation study in \cref{sec:results:ablation}.

\begin{figure}[t]
	\begin{center}
		\begin{subfigure}[c]{0.75\columnwidth}
			\includegraphics[width=1\textwidth]{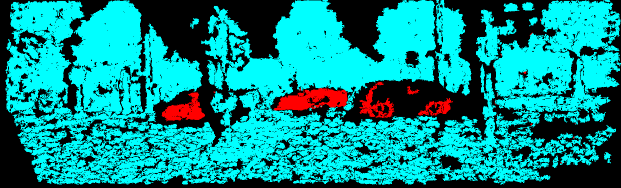}%
			\raisebox{2pt}{\makebox[0pt][r]{\textcolor{white}{\bf a)~}}}%
			\vspace{0.5mm}%
		\end{subfigure}
		\begin{subfigure}[c]{0.75\columnwidth}
			\includegraphics[width=1\textwidth]{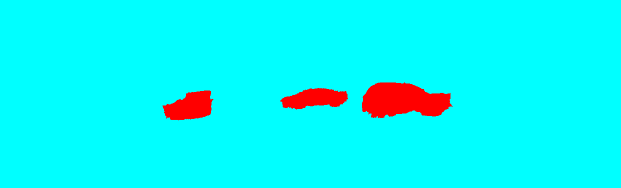}%
			\raisebox{2pt}{\makebox[0pt][r]{\bf b)~}}%
			\vspace{0.5mm}%
		\end{subfigure}
		\begin{subfigure}[c]{0.75\columnwidth}
			\includegraphics[width=1\textwidth]{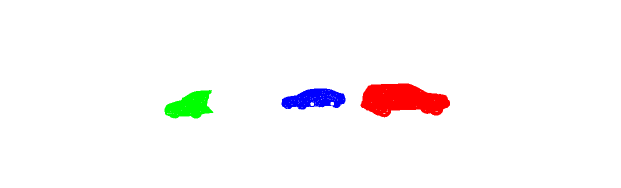}%
			\raisebox{2pt}{\makebox[0pt][r]{\bf c)~}}%
		\end{subfigure}
	\end{center}
	\caption{Example of our motion segmentation. Sparse motion indicators as obtained during ego-motion computation {\bf (a)}, dense segmentation by interpolation {\bf (b)} and moving ground truth objects as provided by KITTI \citep{menze2015object} {\bf (c)}.}
	\label{fig:segmentation}
\end{figure}

\subsection{Optional Ego-Motion Model} \label{sec:egomotion}
In \cref{sec:results} we will show that our approach as described so far achieves competitive results. For the special challenges of the KITTI data set, we make an additional, optional assumption to further improve the performance of \gls*{sff}. Following \citet{taniai2017fsf}, we argue that most parts of a scene are static and thus that the 3D motion for these areas is fully determined by the ego-motion of the observer. Given the ego-motion and a motion segmentation into static and dynamic areas, we apply the inverse ego-motion to all static points in the scene. 
Using our matching and interpolation scheme, both can be estimated easily with almost no additional effort.
However, our improvements in \cref{sec:extension} will make the refinement step by the ego-motion model obsolete. Our experiments will show, that the impact of the ego-motion optimization diminishes when using our more robust interpolation.

\paragraph{Ego-Motion Estimation.}
The filtered scene flow field before interpolation provides very accurate matches across all images. We compute 3D-2D correspondences between the reference frame and the temporally subsequent frame by triangulation with the stereo matches. We limit the depth of these correspondences to $35$ meters because disparity resolution for farther distances becomes too inaccurate. This way, we obtain a Perspective-n-Point (PnP) problem, which we solve iteratively using Levenberg-Marquardt and RANSAC to find the relative pose between the left camera at time $t$ and $t+1$ by minimizing the re-projection error of all correspondences. For RANSAC, we consider a correspondence to be an outlier if the re-projection error is above $1$ pixel. After a first estimation, we recompute the set of inliers with a relaxed threshold of $3$ pixels and re-estimate the pose $P=\left[R|t\right] \in \mathbb{R}^{3 \times 4}$. The two-stage process helps to avoid local optima and to find a trade-off between diverse and robust correspondences.

\paragraph{Motion Segmentation.}
An initial sparse motion segmentation can directly be obtained as a side product of the ego-motion estimation. Outliers in RANSAC are considered \textit{in motion}, while points in conformity with the estimated ego-motion are marked as \textit{static}. We use our boundary-aware interpolation to compute a dense segmentation (cf. \cref{fig:segmentation}). Pixels labeled as moving are spread up to the object boundaries within which they are detected. Because the segmentation is only a binary labeling, no complex interpolation model is needed. An unknown pixel gets assigned with the weighted mean of its local neighborhood. The weights are again based on the geodesic distances between matches. This interpolation method is similar to the Nadaraya-Watson estimator described by \citet{revaud2015epic}. The interpolated motion field is then thresholded to obtain a dense, binary motion segmentation.
Finally, the inverse estimated ego-motion is applied to all points that are labeled as static.
The impact of the ego-motion model is huge for our dual-frame conference approach, as will be shown in \cref{sec:results:ablation}.

\section{Robust Multi-frame Extension} \label{sec:extension}
Our experiments will show that \gls*{sff}, presented in the previous section, yields competitive scene flow results, especially when the ego-motion model is applied. However, there are two main problems with the presented pipeline: 1.) Pixel-wise matching without regularization is error prone under some circumstances (saturation, lighting variations, homogeneous or repetitive textures, etc.) and even impossible in occluded and other invisible image regions (\eg out-of-bounds motions, cf. \cref{fig:visibility:gt}). 2.) The accuracy of the interpolation suffers from increasing gap sizes in the filtered scene flow field and from remaining outliers after consistency check. The ego-motion model can compensate inaccurate interpolation to some extend, but not sufficiently.
In this paper, we address both major issues with our novel robust extension to multiple frame pairs.

\begin{table}[t]
	\centering
	\caption{A comparison between the dual- and multi-frame pipeline. We contrast the individual steps of the pipeline. The differences are visualized in \cref{fig:overview}.}
	\label{tab:contrast}
	\resizebox{1\linewidth}{!}{
	\begin{tabular}{c|c|c}
		& \begin{tabular}{c}\bf \gls*{sff}\\dual-frame\end{tabular} & \begin{tabular}{c} \bf \gls*{sffpp}\\multi-frame \end{tabular}\Bstrut\\
		\hline
		Initialization & kD-Trees & Previous result\Tstrut\\
		Matching & Two time steps & \begin{tabular}{c} Three time steps\\and visibility reasoning \end{tabular}\\
		\begin{tabular}{c} Consistency\\Check \end{tabular} & Fully inverse & Left-right\\
		Interpolation & Edge-preserving & \begin{tabular}{c} Edge-preserving\\and robust \end{tabular}\\
		Refinement & \begin{tabular}{c} Variational\\and ego-motion \end{tabular} & Not needed\\	
	\end{tabular}
	}
\end{table}

There are different techniques to handle the unmatchable parts of a scene, \ie cases where it is difficult or even impossible to find correspondences in the relevant images. Typically, some kind of regularization is applied, like in form of a smoothness assumption which encodes that neighboring pixels should represent a similar motion so that local visual evidence can support the motion estimation in the difficult areas. In methods that employ the piece-wise rigid plane model \citep{vogel2013PRSF,menze2015object,vogel2015PRSM,lv2016CSF,neoral2017object,behl2017bounding} this kind of regularization is for two reasons considerably strong. Firstly, each local patch describing a slanted plane undergoes the same transformation by design. Secondly, inter-plane smoothness is further enforced by dedicated terms in the energy formulation. However, regularization terms increase the computational effort significantly (cf. \cref{tab:kitti}), and would prohibit the use of our efficient optimization strategy of \cref{eq:energy}.
An alternate concept to handle unmatchable regions is sparse-to-dense interpolation. This idea is rather young and was first successfully realized by EPICFlow \citep{revaud2015epic} for the optical flow problem. The idea is to remove regions of low confidence (\ie regions where regularization would be required to match them accurately) and to use interpolation to fill the gaps based on reliable matches.
\gls*{sff} is to the best of our knowledge the first method of sparse-to-dense interpolation for scene flow correspondences and our extension is transferring this concept to a multi-frame setup for the first time.
Even though the consistency check removes outliers reliably, the gaps can not be refilled by the interpolation correctly in some scenarios. Therefore, we tackle the problem before it occurs by using image information from multiple frames to avoid mismatches and resolve ambiguity in unmatchable regions. This will result in more accurate and better distributed matches of higher density as will be shown in \cref{sec:results:extension}. Further, we transfer the concepts for interpolation from RICFlow \citep{hu2016efficient} to the scene flow domain to improve robustness during interpolation.

All our extensions using more than two stereo image pairs build on the assumption of constant motion. That means, we assume the observed motion at both time steps (from $t-1$ to $t$ and from $t$ to $t+1$) to be the same. The error made by this assumption converges towards zero for continuous motions with increasing frame rates.
Our multi-frame approach is additionally designed to process video streams in an online manner. That means that the first two frame pairs of a sequence are processed with the dual-frame approach from \cref{sec:sff}. All subsequent frames are than added in an incremental way and processed within a sliding temporal window of three frame pairs.
We exploit the additional information of the extra images during matching in three ways. First, the previous results are used during initialization. Secondly, we use the previous scene flow to predict visibility of the scene. Lastly, we use our assumption together with the visibility prediction to match scene flow across all six relevant images.

The overall structure of our multi-frame approach remains the same as in the dual-frame method but with two more images. Using a set of six input images $\mathbf{I} = \{I_0^l, I_0^r, I_1^l, I_1^r, I_{-1}^l, I_{-1}^r\}$, we find accurate matches, remove possible outliers, and interpolate back to a dense scene flow field. The overview of the multi-frame pipeline is shown in \cref{fig:overview}. A comparison between \gls*{sff} and  \gls*{sffpp} is given in \cref{tab:contrast}.

\subsection{Multi-frame Matching} \label{sec:extension:matching}

\begin{figure}[t]
	\centering
	\begin{subfigure}[c]{0.475\columnwidth}
		\includegraphics[width=1\textwidth]{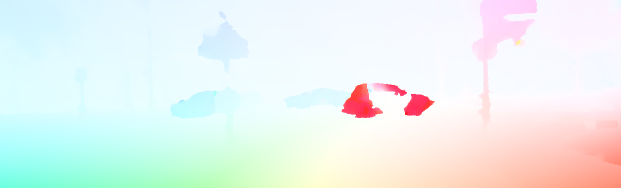}%
		\vspace{0.5mm}%
	\end{subfigure}
	\begin{subfigure}[c]{0.475\columnwidth}
		\includegraphics[width=1\textwidth]{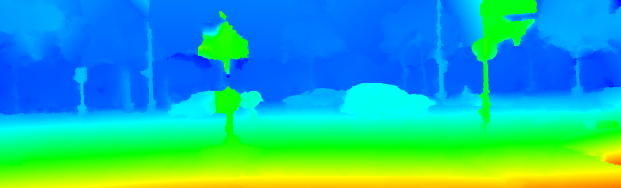}%
		\raisebox{2pt}{\makebox[0pt][r]{\bf a)~}}%
		\vspace{0.5mm}%
	\end{subfigure}
	\begin{subfigure}[c]{0.475\columnwidth}
		\includegraphics[width=1\textwidth]{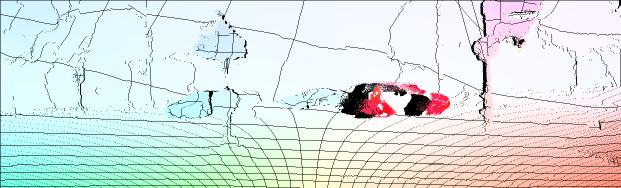}%
		\vspace{0.5mm}%
	\end{subfigure}
	\begin{subfigure}[c]{0.475\columnwidth}
		\includegraphics[width=1\textwidth]{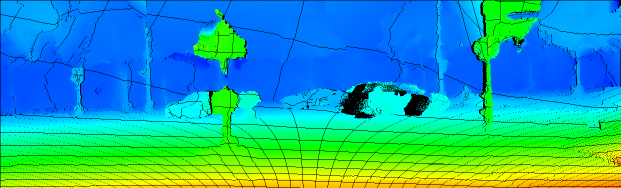}%
		\raisebox{2pt}{\makebox[0pt][r]{\bf b)~}}%
		\vspace{0.5mm}%
	\end{subfigure}
	\begin{subfigure}[c]{0.475\columnwidth}
		\includegraphics[width=1\textwidth]{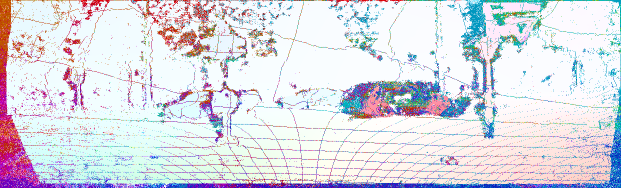}
	\end{subfigure}
	\begin{subfigure}[c]{0.475\columnwidth}
		\includegraphics[width=1\textwidth]{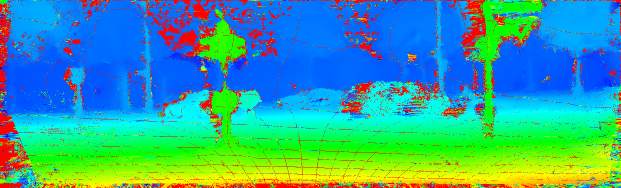}%
		\raisebox{2pt}{\makebox[0pt][r]{\bf c)~}}%
	\end{subfigure}
	\caption{Previous scene flow result \textbf{(a)}, temporally warped scene flow \textbf{(b)}, and multi-frame initialization \textbf{(c)} (on full resolution for visualization purposes).}
	\label{fig:multi_init}
\end{figure}

\paragraph{Initialization.}
Improved initialization is the first extension in our multi-frame approach. We use the previous scene flow result and apply our assumption of constant motion. We propagate each 3D point according to its 3D motion to get an initial prediction of the scene flow at the current time step. We use the temporally propagated scene flow prediction as additional candidate during the initialization process earlier described in \cref{sec:sff:matching}. The process is visualized in \cref{fig:multi_init}. By not relying on the previous result alone, we successfully avoid error propagation (cf. \cref{fig:multi_init}).

\paragraph{Matching Cost.}
As before, our matching process is based on the visual similarity of corresponding pixels (cf. \cref{eq:matchingcost}). We add the additional image correspondences to our scene flow matching cost from \cref{eq:cost_dual} and obtain
\begin{equation} \label{eq:cost_multi}
\begin{split}
	& C_{multi}\left(\mathbf{p},\mathbf{u}\right) = C\left(I_0^l,\mathbf{p},I_1^l,\mathbf{p} +\left(u,v\right)^T\right)\\
	& + C\left(I_0^l,\mathbf{p},I_0^r,\mathbf{p}+\left(-d_0,0\right)^T\right)\\
	& + C\left(I_0^l,\mathbf{p},I_1^r,\mathbf{p}+\left(u-d_1,v\right)^T\right)\\
	& + C\left(I_0^l,\mathbf{p},I_{-1}^l,\mathbf{p}+\left(u_{-1},v_{-1}\right)^T\right)\\
	& + C\left(I_0^l,\mathbf{p},I_{-1}^r,\mathbf{p}+\left(u_{-1}-d_{-1},v_{-1}\right)^T\right).
\end{split}
\end{equation}
We match all pixels of the image domain $\Omega$ to all five other images and derive corresponding pixel locations from our scene flow representation in 2D (cf. \cref{eq:cost_multi}). For matching with the previous frame pair, we compute the inverse scene flow (in image space) $u_{-1}$, $v_{-1}$, $d_{-1}$ by projecting the flow to 3D, inverting it, and projecting back to 2D, according to our constant motion assumption. 
Note that $u_{-1} = -u$ if and only if $d_0 = d_1$. We do not simply invert the 2D optical flow directly. The (pixel-wise) constant motion assumption allows us to match across multiple time steps without increasing the search space or complexity of the scene flow domain.

\begin{figure*}[t]
	\centering
	\begin{subfigure}[b]{0.98\linewidth}
		\centering
		\includegraphics[width=0.3\columnwidth]{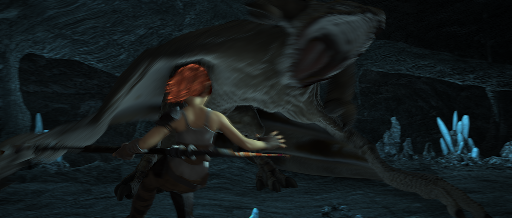}
		\includegraphics[width=0.3\columnwidth]{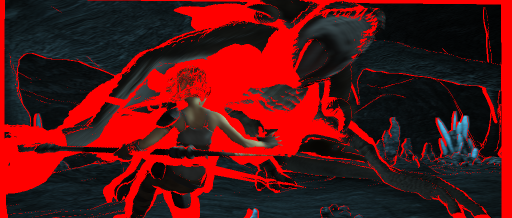}
		\includegraphics[width=0.3\columnwidth]{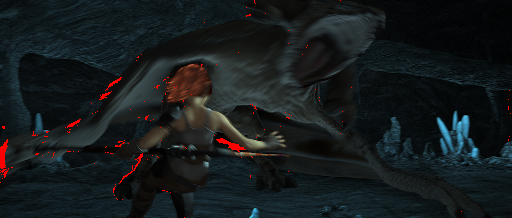}
		\vspace{1mm}%
		\caption{Invisibility Example. Reference image (left), invisibility mask in the dual-frame scenario where no visibility handling is applied (middle), and remaining unmatchable areas due to occlusions in the multi-frame scenario where visibility handling is applied (right).} \label{fig:visibility:gt}
		\vspace{1mm}%
	\end{subfigure}
	\begin{subfigure}[b]{0.95\linewidth}
		\centering
		\hspace{0.05\columnwidth}%
		\begin{subfigure}[c]{0.44\columnwidth}
		\centering Left View
		\end{subfigure}
		\begin{subfigure}[c]{0.44\columnwidth}
		\centering Right View
		\end{subfigure}
		\begin{subfigure}[c]{0.05\columnwidth}
			\Bstrut
		\end{subfigure}\\%
		\hspace{0.05\columnwidth}%
		\begin{subfigure}[c]{0.44\columnwidth}
			\includegraphics[width=1\columnwidth]{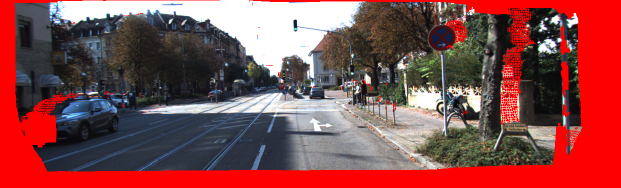}%
			\vspace{1mm}%
		\end{subfigure}
		\begin{subfigure}[c]{0.44\columnwidth}
			\includegraphics[width=1\columnwidth]{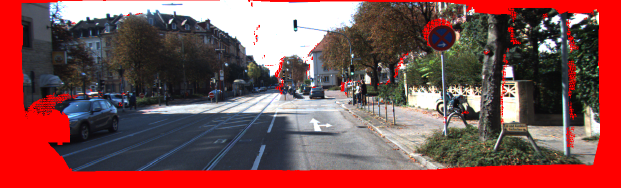}%
			\vspace{1mm}%
		\end{subfigure}
		\begin{subfigure}[c]{0.05\columnwidth}
			\rotatebox[origin=c]{90}{$t+1$}
		\end{subfigure}\\%
		\hspace{0.05\columnwidth}%
		\begin{subfigure}[c]{0.44\columnwidth}
			\includegraphics[width=1\columnwidth]{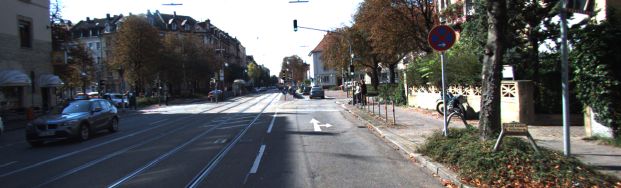}%
			\vspace{1mm}%
		\end{subfigure}
		\begin{subfigure}[c]{0.44\columnwidth}
			\includegraphics[width=1\columnwidth]{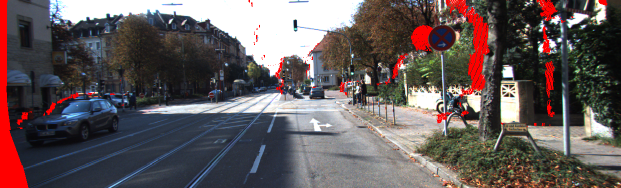}%
			\vspace{1mm}%
		\end{subfigure}
		\begin{subfigure}[c]{0.05\columnwidth}
			\rotatebox[origin=c]{90}{$t$}
		\end{subfigure}\\%
		\hspace{0.05\columnwidth}%
		\begin{subfigure}[c]{0.44\columnwidth}
			\includegraphics[width=1\columnwidth]{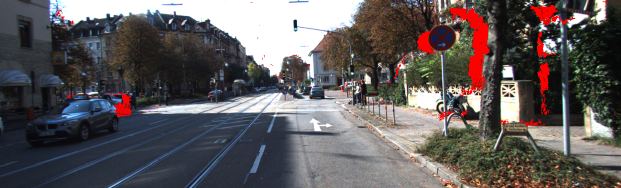}%
		\end{subfigure}
		\begin{subfigure}[c]{0.44\columnwidth}
			\includegraphics[width=1\columnwidth]{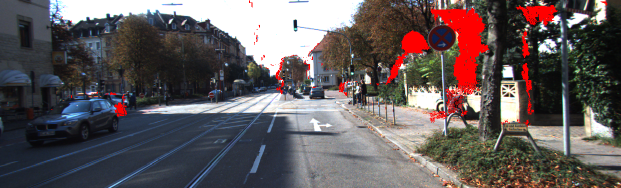}%
		\end{subfigure}
		\begin{subfigure}[c]{0.05\columnwidth}
			\rotatebox[origin=c]{90}{$t-1$}
		\end{subfigure}
		\caption{Visibility Prediction} \label{fig:visibility:predict}
	\end{subfigure}
	\caption{In the dual-frame case, occlusions can obscure large parts of the image which remain mostly visible when using multiple frames (\subref{fig:visibility:gt}). Our explicit visibility handling of the multi-frame matching strategy predicts occluded and out-of-bounds regions (red) for all five images that are matched to the reference frame $I_0^l$ (\subref{fig:visibility:predict}).} \label{fig:visibility}
\end{figure*}

\paragraph{Visibility Prediction.}
One additional major extension is our explicit visibility reasoning. Since the energy is based on visual data only, it is impossible to match regions, that are not visible in one of the images. Depending on the magnitude of camera movement, the baseline, and other circumstances, those unmatchable areas can become considerably large. \cref{fig:visibility:gt} shows an invisibility mask for pixels whose imaged 3D point is not visible in at least one of the views of $I_0^r$, $I_1^l$, or $I_1^r$. This gives an impression of the limitations of dual-frame matching methods. 
However, \cref{fig:visibility:gt} also visualizes the remaining invisibility when considering one additional time step.
More than two frame pairs can compensate for missing visual evidence when a pixel correspondence that is invisible in one image can be observed in another. Assuming to know which pixels are occluded or out-of-bounds of the image domain in each view, we can replace the matching cost for a single image correspondence (\cref{eq:matchingcost}) in \cref{eq:cost_multi} by
\begin{multline} \label{eq:cost_vis}
	C_{\text{vis}}\left(I,\mathbf{p},I',\mathbf{p'}, {occ'}, {oob'}\right) = \\
	\begin{cases}
	\theta_{occ}, & \text{if } occ'\left(\mathbf{p}\right) \\
	\theta_{oob}, & \text{if } oob'\left(\mathbf{p}\right) \land \mathbf{p'} \notin \Omega \\
	\theta_{penalty}, & \text{if } oob'\left(\mathbf{p}\right) \neq \mathbf{p'} \in \Omega \\
	C\left(I,\mathbf{p},I',\mathbf{p'}\right), & \text{else,} \\	
	\end{cases}
\end{multline}
with $occ'$ and $oob'$ being binary occlusion and out-of-bounds masks that indicate for each pixel $\mathbf{p}$ whether the corresponding point in view $I'$ is visible. For our full multi-frame matching cost in \cref{eq:cost_multi}, we need two invisibility masks for all 5 corresponding images. The adjusted cost uses a constant invisibility cost $\theta_{occ}$ in case of occlusions to avoid trivial solutions where all points would be occluded. Same holds for out-of-bounds motions, with the additional option to assign a very large penalty cost $\theta_{penalty}$ if motion under test is inconsistent with the out-of-bounds mask $oob'$. Otherwise, the normal matching cost of \cref{eq:matchingcost} will be used. In practice, we choose $\theta_{occ} = \theta_{oob} = 10\,000$ based on empirical studies of the matching cost of ground truth scene flow vectors and make $\theta_{penalty}=10^6$ large enough so that out-of-bounds motion is forced to be consistent with the prediction.

Because the algorithm is designed to process sequences sequentially, we can use the estimated scene flow of the previous time step to predict the visibility for all six images. To this end, we use the temporally propagated scene flow prediction as for the initialization. This propagation can be used to check which parts of the scene leave the image domain $\Omega$ for a specific view. These areas are marked as invisible in the associated out-of-bounds mask. It also allows to reason about occlusions by z-buffering, since we have full 3D information including depth. If multiple motions have the same target pixel in the target view, all but the closest are occluded. Examples of the visibility prediction are given in \cref{fig:visibility:predict}. For each of the five relevant frames, we show which pixel position of the reference frame can not be observed from the respective view. All pixels of the reference frame are visible by definition.

The impact of our multi-frame strategy combined with the explicit visibility reasoning will be shown in \cref{sec:results:extension} by comparing to the basic dual-frame \gls*{sff} from \cref{sec:sff} \citep{schuster2018sceneflowfields}.

\paragraph{Consistency Check.}
The next step of the pipeline is the consistency check to remove possible outliers. As before, we do so by computing a consistency scene flow field and compare corresponding scene flow vectors. 
Due to the changed setup to multiple frames from a stereo sequence, the way the consistency field is computed needs to be changed. We do not invert the temporal order of the images (cf. \cref{sec:sff:matching}), but only change the view point. This leaves us with a left-right consistency check. Although a left-right check alone is less reliable than our previous consistency check, this has an important advantage. With inverted temporal order, scene flow correspondences according to the optical flow can not be established where the motion leaves the image boundaries. Thus, out-of-bounds regions are always filtered. This was no issue in the dual-frame approach, but using multi-frames we are able to match scene flow in these regions. With the left-right consistency check, we have the ability to maintain the correct correspondences in out-of-bounds regions. 
Matching for the consistency scene flow field is done exactly as before, using multiple scales, visibility reasoning, etc. Afterwards, each scene flow estimate of the reference frame is compared to its corresponding vector of the consistency field component-wise. If any error exceeds $\tau_c=1$ pixel, the whole scene flow vector is removed. 
\gls*{sff} did benefit from an additional consistency check for the disparity only to obtain more matches for geometry interpolation. We follow that idea and compare our dense multi-frame matches to a second independent disparity map computed with the algorithm from \citet{yamaguchi2014efficient}. Afterwards, we take the union of the additional matches and the results of the previous consistency check. 
We did change the stereo algorithm from \gls*{sgm} \citep{hirschmuller2008SGM} to SPS \citep{yamaguchi2014efficient} because it is faster, more accurate, and also able to match occluded and other invisible regions as our matching approach.
The remaining merged correspondences are very accurate and serve as input for the robust interpolation (cf. \cref{fig:overview,fig:dual_vs_multi}).

\subsection{Robust Interpolation} \label{sec:extension:interpolation}
We apply the same sparsification as in \gls*{sff} because sparse-to-dense interpolation works best if input matches are not too dense already  \citep{schuster2018sceneflowfields,bailer2015flow,hu2016efficient}. The sparsification works as an additional outlier filter and increases the spatial support for the same number of neighbors during interpolation.

\begin{figure}[t]
	\centering
	\begin{subfigure}[c]{0.9\columnwidth}
		\adjincludegraphics[width=1\columnwidth,trim={0.2\width} {0} {0.35\width} {0.6\height},clip]{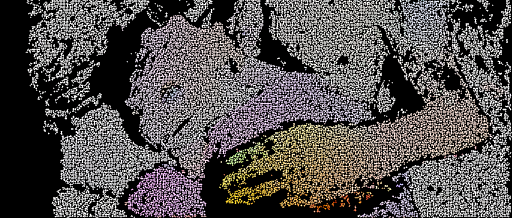}%
		\raisebox{2pt}{\makebox[0pt][r]{\textcolor{white}{\bf a)~}}}%
		\vspace{1mm}%
	\end{subfigure}
	\begin{subfigure}[c]{0.9\columnwidth}
		\adjincludegraphics[width=1\columnwidth,trim={0.2\width} {0} {0.35\width} {0.6\height},clip]{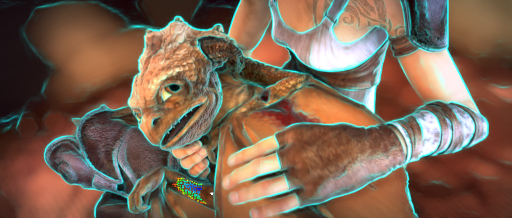}%
		\raisebox{2pt}{\makebox[0pt][r]{\textcolor{white}{\bf b)~}}}%
	\end{subfigure}
	\caption{Sparsified input matches for interpolation \textbf{(a)} and the edge-aware neighborhood of seeds for the superpixel in white \textbf{(b)}. Note that this is the actual size of segments on which we apply our interpolation models on. The supporting seeds are colored according to their distance to the superpixel.} \label{fig:interpolation}
\end{figure}

The interpolation in our extension is robust against possible remaining outliers and edge-preserving to create sharp motion boundaries at object edges. As in the less robust approach, we split interpolation into interpolation of the 3D geometry, followed by interpolation of the 3D motion. Because of this separation, we can use different sets of input seeds for each. Other than SFF in \cref{sec:sff} \citep{schuster2018sceneflowfields} where the interpolation algorithm of \citet{revaud2015epic} was transferred to the higher dimensional scene flow problem, we adopt the robust concepts of \citet{hu2017robust} for our scene flow problem. 

In short, we dissect the reference frame into superpixels of size 25. Each superpixel is associated with a local neighborhood of the 200 closest input matches as shown in \cref{fig:interpolation}. The distance is computed as a geodesic distance based on an edge map. Additionally, each superpixel is initialized with two scene flow models, one for the 3D geometry and one for the 3D motion. Afterwards, the models for each superpixel is adjusted by propagation and random search. Based on the final models, dense scene flow can be computed for each pixel.

\paragraph{Interpolation Models.}
The geometric model is a slanted plane for each superpixel. It is initialized with a constant depth, \ie parallel to the image plane. 
The motion model is a rigid transformation of 3D rotation and 3D translation for each superpixel. This rigid model is less universal, but more robust compared to our affine model in \cref{sec:sff:interpolation}. We do not assume rigid motions, though. Our approach is not limited to rigid motions. Due to the small size of the superpixel segmentation, we can approximate non-rigid motions closely. Same holds for the planarity. Arbitrarily curved surfaces will be approximated by very small plane segments. The superpixel motion is initialized with translation only.
For initialization, three different strategies have been tried: 1.) The basic approach of \citet{hu2017robust} where the initial values are obtained from the the closest input seed. 2.) One where the local weighted medoid is used. 3.) One estimating the weighted geometric median. Though the multivariate estimators are much more accurate than the nearest neighbor method, the robust propagation with random search can compensate for less accurate initialization. Independent of the superpixel initialization, the final interpolation result was almost the same in all cases. We decided to use the robust geometric median in practice, since it provides a reasonable tradeoff between accuracy and run time.

\paragraph{Robust Model Estimation.}
To optimize the models for interpolation (plane + rigid transformation), we apply a robust, stochastic approach in a RANSAC-like fashion.
For each superpixel, we randomly sample seeds from a local edge-aware neighborhood and use those to predict the models. Apart from random sampling, we use propagation to test estimated models at neighboring superpixels.
To determine the fitness of a model, a truncated error for all supporting input seeds $\mathbf{s}$ of the local neighborhood $\mathcal{N}$ gets summed
\begin{equation}
	C\left(M\right) = \sum_{\mathbf{s} \in \mathcal{N}} \text{min}\left(\tau, \exp{\left(-\frac{1}{\alpha} \cdot D\left(\mathbf{s}\right) \right)} \cdot \epsilon\left(\mathbf{s}\right) \right).
\end{equation}
In this formula, $M$ is the respective model under test, $\tau$ is the truncation error threshold of 4 pixels, and $\epsilon\left(\mathbf{s}\right)$ is the re-projection error when applying model $M$ to input seed $\mathbf{s}$ which is weighted by a geodesic distance $D$ between the seed $\mathbf{s}$ and the position of the respective superpixel with weight coefficient $\alpha=0.6$. By minimizing the cost of all models for all superpixels using propagation and random model generation, we obtain robust interpolation models for geometry and motion.
As before with multi-frame matching, our novel robust interpolation is compared directly to the previous interpolation method for scene flow matches of \cref{sec:sff:interpolation} \citep{schuster2018sceneflowfields} in \cref{sec:results:extension}.

\begin{figure}[t]
	\centering
	\begin{subfigure}[c]{0.48\linewidth}
		\centering
		\includegraphics[width=1\linewidth]{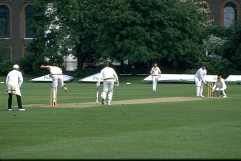}%
		\vspace{1mm}%
	\end{subfigure}
	\begin{subfigure}[c]{0.48\linewidth}
		\centering
		\includegraphics[width=1\linewidth]{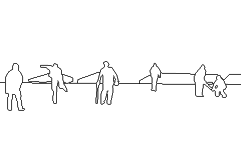}%
		\raisebox{2pt}{\makebox[0pt][r]{\bf a)~}}%
		\vspace{1mm}
	\end{subfigure}\\%
	\begin{subfigure}[c]{0.48\linewidth}
		\centering
		\includegraphics[width=1\linewidth]{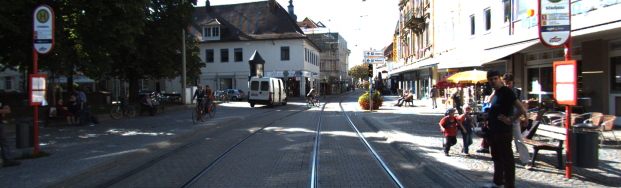}%
		\vspace{1mm}%
	\end{subfigure}
	\begin{subfigure}[c]{0.48\linewidth}
		\centering
		\includegraphics[width=1\linewidth]{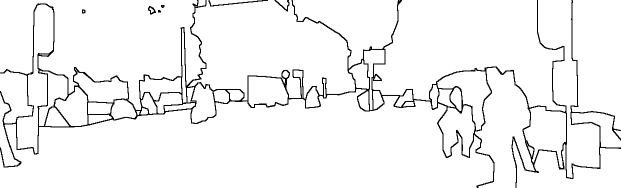}%
		\raisebox{2pt}{\makebox[0pt][r]{\bf b)~}}%
		\vspace{1mm}%
	\end{subfigure}\\%
	\begin{subfigure}[c]{0.48\linewidth}
		\centering
		\includegraphics[width=1\linewidth]{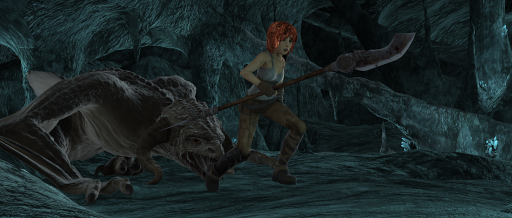}%
	\end{subfigure}
	\begin{subfigure}[c]{0.48\linewidth}
		\centering
		\includegraphics[width=1\linewidth]{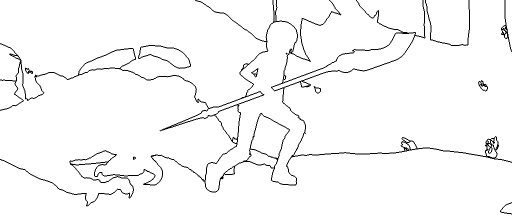}%
		\raisebox{2pt}{\makebox[0pt][r]{\bf c)~}}%
	\end{subfigure}
	\caption{Examples of boundaries from BSDS \citep{martin2001bsds} \textbf{(a)}, KITTI \citep{geiger2012kitti} \textbf{(b)}, and Sintel \citep{butler2012sintel} \textbf{(c)} used to train our universal edge detector.} 
	\label{fig:gtedges}
\end{figure}

\paragraph{Universal Boundary Detector.}
We will show that the boundary detector from \cref{sec:sff:interpolation}, trained on semantic boundaries of KITTI, performs much better than the original detector \citep{dollar2013sed} on that particular domain. However, improved robustness requires reliability across multiple domains. Thus, we train a third variant with the goal to perform equally well on several different data sets.
Towards this end, we use a joint set of training images from the 200 images of the original BSDS data set \citep{martin2001bsds} and add semantic boundaries for 424 images of KITTI and for 100 images of Sintel. The semantics for KITTI are the same as before. For Sintel, we use the sequences \textit{cave\_2} and \textit{sleeping\_1} of the \textit{clean} rendering pass that are excluded during evaluation. We create semantics by merging the provided mesh and material segmentation, e.g. all segments belonging to the dragon are labeled with the same ID. Examples of ground truth boundaries for each of the data sets are given in \cref{fig:gtedges}. By combining these three sets, we obtain a total of 724 images for training. The combined edge model is much more versatile and applicable to different data sets.
Examples of the detected edges are visualized in \cref{fig:sintel,fig:title:image,fig:edges,fig:interpolation}. Textures and shadows are suppressed, while at the same time, object boundaries are detected accurately.

For optional refinement of the dense 3D motion field, we use variational optimization of \cref{sec:variational} and the ego-motion model of \cref{sec:egomotion}. However, our improvements in matching and interpolation make both refinement steps obsolete as we will show in \cref{sec:results:ablation}.

\section{Experiments and Results} \label{sec:results}
A series of experiments was conducted to evaluate our scene flow algorithm. We start with an ablation study to show the impact of each individual step of our sparse-to-dense pipeline and the incremental improvements presented within this paper. It follows a more detailed comparison between the dual-frame approach (\cref{sec:sff}) and the robust multi-frame extension (\cref{sec:extension}). Afterwards, we will present results on the public KITTI scene flow benchmark and on the MPI Sintel dataset for comparison to state-of-the art.
Finally, we disclose the limitations of our approach.

For all our experiments, we keep the parameters of \gls*{sffpp} fixed at the given values in the previous sections. For \gls*{sff}, we use the parameters of the original approach \citep{schuster2018sceneflowfields} which are not identical for KITTI and Sintel. The other approaches in our comparison also use adjusted parameters for the Sintel data set.

\begin{table}[t]
	\centering
	\caption{Results of our scene flow estimation pipeline for different combinations of matching and interpolation settings, before and after variational optimization and ego-motion refinement.}
	\label{tab:ablation}
	\resizebox{1\linewidth}{!}{
	\begin{tabular}{l||c|c|c}
		\textbf{Method} & \textbf{SF-bg} & \textbf{SF-fg} & \textbf{SF-all}\Bstrut\\
		\hline
		dual+epic3d+bsds & 26.93 & 32.16 & 27.73\Tstrut\\
		dual+epic3d+bsds+var & 26.14 & 30.71 & 26.84 \\
		dual+epic3d+bsds+var+ego & 13.39 & 30.92 & 16.07\Bstrut\\
		\hline
		dual+epic3d+kitti & 25.60 & 28.99 & 26.12\Tstrut\\
		dual+epic3d+kitti+var & 24.78 & 27.37 & 25.18 \\
		dual+epic3d+kitti+var+ego (SFF) & 12.04 & 28.31 & 14.53\Bstrut\\
		\hline
		dual+ric3d+kitti & 12.10 & 22.86 & 13.74\Tstrut\\
		dual+ric3d+kitti+var & 12.11 & 22.96 & 13.77 \\
		dual+ric3d+kitti+var+ego & 11.57 & 23.46 & 13.40\Bstrut\\
		\hline
		multi+ric3d+kitti & 10.75 & 19.37 & 12.07\Tstrut\\
		multi+ric3d+kitti+var & 11.26 & 19.60 & 12.54 \\
		multi+ric3d+kitti+var+ego & 11.27 & 20.11 & 12.62\Bstrut\\
		\hline
		multi+ric3d+mixed (SFF++) & 10.93 & 19.67 & 12.27\Tstrut\\
		multi+ric3d+mixed+var & 11.42 & 20.12 & 12.75 \\
		multi+ric3d+mixed+var+ego & 11.29 & 20.62 & 12.72 \\
	\end{tabular}
	}
\end{table}

\begin{table*}[p]
	\centering
	\caption{Comparison of intermediate results for \gls*{sff} \citep{schuster2018sceneflowfields} and our novel robust multi-frame extension \gls*{sffpp} on the KITTI training data.} 
	\label{tab:dual_vs_multi}
	\resizebox{1\textwidth}{!}{\begin{tabular}{c c || c | c | c || c | c | c | c || c || c || c | c | c | c || c | c | c }
		\multicolumn{2}{c||}{} & \multicolumn{8}{c||}{\textit{All pixels}} & \multicolumn{8}{c}{\textit{Occluded pixels only}}\\
		\multicolumn{2}{c||}{} & \multicolumn{3}{c||}{\bf EPE [px]} & \multicolumn{4}{c||}{\bf Outliers [\%]} & \multicolumn{2}{c||}{\bf Density [\%]} & \multicolumn{4}{c||}{\bf Outliers [\%]} & \multicolumn{3}{c}{\bf EPE [px]}\\
		\multicolumn{2}{c||}{} & \textbf{D1} & \textbf{D2} & \textbf{Fl} & \textbf{D1} & \textbf{D2} & \textbf{Fl} & \textbf{SF} & \multicolumn{2}{c||}{\bf SF} & \textbf{D1} & \textbf{D2} & \textbf{Fl} & \textbf{SF} & \textbf{D1} & \textbf{D2} & \textbf{Fl}\Bstrut\\
		\hline
		\multirow{2}{*}{Matching} & SFF & 7.2 & 11.3 & 38.9 & 12.6 & 29.2 & 32.8 & 39.8 & 100.0 & 100.0 & 22.1 & 92.1 & 95.5 & 99.5 & 34.3 & 50.1 & 195.7\Tstrut\\
		& \textbf{SFF++} & \textbf{2.7} & \textbf{4.3} & \textbf{9.2} & \textbf{11.4} & \textbf{20.9} & \textbf{23.7} & \textbf{31.8} & 100.0 & 100.0 & \textbf{14.9} & \textbf{39.8} & \textbf{55.3} & \textbf{62.1} & \textbf{5.3} & \textbf{7.5} & \textbf{25.9}\Bstrut\\
		\hline
		\multirow{2}{*}{Filtered} & SFF & \textbf{0.8} & \textbf{0.9} & \textbf{1.1} & \textbf{2.2} & \textbf{2.7} & \textbf{2.2} & \textbf{4.2} & 38.8 & 0.3 & 19.5 & 76.0 & 75.2 & 79.0 & 4.6 & 13.3 & 79.9\Tstrut\\
		& \textbf{SFF++} & 0.9 & 1.1 & 1.6 & 2.6 & 4.7 & 5.3 & 8.0 & \textbf{41.6} & \textbf{9.8} & \textbf{2.5} & \textbf{12.9} & \textbf{23.0} & \textbf{26.2} & \textbf{1.1} & \textbf{2.4} & \textbf{9.3}\Bstrut\\
		\hline
		\multirow{2}{*}{Interpolated} & SFF & \textbf{1.2} & 3.7 & 16.1 & 6.2 & 16.2 & 23.8 & 26.1 & 100.0 & 100.0 & 9.9 & 43.6 & 62.4 & 64.2 & 2.0 & 11.4 & 86.1\Tstrut\\
		& \textbf{SFF++} & \textbf{1.2} & \textbf{1.9} & \textbf{4.7} & \textbf{5.2} & \textbf{8.8} & \textbf{9.7} & \textbf{12.7} & 100.0 & 100.0 & \textbf{7.1} & \textbf{16.0} & \textbf{22.7} & \textbf{24.8} & \textbf{1.7} & \textbf{3.3} & \textbf{15.3}\\
	\end{tabular}}
\end{table*}

\begin{table*}[p]
	\caption{Results on the KITTI Scene Flow Benchmark \citep{menze2015object}.}
	\label{tab:kitti}
	\centering
		\resizebox{0.95\textwidth}{!}{\begin{tabular}{c || c | c | c || c | c | c || c | c | c || c | c | c || c}
 & \multicolumn{3}{c||}{\bf D1} & \multicolumn{3}{c||}{\bf D2} & \multicolumn{3}{c||}{\bf Fl} & \multicolumn{3}{c||}{\bf SF} & {\bf Run}\\
{\bf Method} & {\bf bg} & {\bf fg} & {\bf all} & {\bf bg} & {\bf fg} & {\bf all} & {\bf bg} & {\bf fg} & {\bf all} & {\bf bg} & {\bf fg} & {\bf all} & {\bf Time} \Bstrut\\ 
\hline
\begin{tabular}{c}ISF\\\citep{behl2017bounding}\end{tabular} & 4.12 & \textbf{6.17} & 4.46 & \textbf{4.88} & \textbf{11.34} & \textbf{5.95} & 5.40 & \textbf{10.29} & \textbf{6.22} & \textbf{6.58} & \textbf{15.63} & \textbf{8.08} & 600 s\Tstrut\\
\begin{tabular}{c}PRSM\\\citep{vogel2015PRSM}\end{tabular} & \textbf{3.02} & 10.52 & \textbf{4.27} & 5.13 & 15.11 & 6.79 & \textbf{5.33} & 13.40 & 6.68 & 6.61 & 20.79 & 8.97 & 300 s \\
\begin{tabular}{c}OSF+TC\\\citep{neoral2017object}\end{tabular} & 4.11  & 9.64 & 5.03  & 5.18 & 15.12 & 6.84 & 5.76 & 13.31& 7.02 & 7.08 & 20.03 & 9.23 & 3000 s\\
\begin{tabular}{c}OSF18\\\citep{menze2018osf}\end{tabular} & 4.11 & 11.12 & 5.28 & 5.01 & 17.28 & 7.06 & 5.38 & 17.61 & 7.41 & 6.68 & 24.59 & 9.66 & 390 s \\ 
\begin{tabular}{c}SSF\\\citep{ren2017cascaded}\end{tabular} & 3.55 & 8.75 & 4.42 & 4.94 & 17.48 & 7.02 & 5.63 & 14.71 & 7.14 & 7.18 & 24.58 & 10.07 & 300 s \\
\begin{tabular}{c}OSF\\\citep{menze2015object}\end{tabular} & 4.54  & 12.03  & 5.79 & 5.45 & 19.41 & 7.77  & 5.62  & 18.92  & 7.83  & 7.01  & 26.34  & 10.23  & 3000 s\\
\begin{tabular}{c}\textbf{SFF++}\\\textbf{(ours)}\end{tabular} & 4.27 & 12.38 & 5.62 & 7.31 & 18.12 & 9.11 & 10.63 & 17.48 & 11.77 & 12.44 & 25.33 & 14.59 & 78 s \\
\begin{tabular}{c}FSF+MS\\\citep{taniai2017fsf}\end{tabular} & 5.72  & 11.84  & 6.74  & 7.57  & 21.28  & 9.85  & 8.48  & 25.43  & 11.30  & 11.17  & 33.91  & 14.96  & {\bf 2.7 s} \\
\begin{tabular}{c}CSF\\\citep{lv2016CSF}\end{tabular} & 4.57  & 13.04  & 5.98  & 7.92  & 20.76  & 10.06  & 10.40  & 25.78  & 12.96  & 12.21  & 33.21  & 15.71  & 80 s \\
\begin{tabular}{c}SFF\\\citep{schuster2018sceneflowfields}\end{tabular} & 5.12 & 13.83 & 6.57 & 8.47 & 21.83 & 10.69 & 10.58 & 24.41 & 12.88 & 12.48 & 32.28 & 15.78 & 65 s \\
\begin{tabular}{c}PRSF\\\citep{vogel2013PRSF}\end{tabular} & 4.74  & 13.74  & 6.24  & 11.14  & 20.47  & 12.69  & 11.73  & 24.33  & 13.83  & 13.49  & 31.22  & 16.44  & 150 s \\
	\end{tabular}}
\end{table*}

\begin{table*}[p]
	\centering
	\caption{Results on MPI Sintel \citep{butler2012sintel}. We give average outliers (EPE \textgreater 3 px) for disparity and optical flow on each sequence separately and averaged over all sequences. We compare PRSM \citep{vogel2015PRSM}, OSF \citep{menze2015object}, FSF \citep{taniai2017fsf}, SFF \citep{schuster2018sceneflowfields}, and our novel SFF++.}
	\label{tab:sintel}
		\begin{tabular}{c||c|c|c|c|c||c|c|c|c|c}
		  & \multicolumn{5}{c||}{{\bf Disparity}} & \multicolumn{5}{c}{{\bf Optical Flow}} \\
		 {\bf Sequence} & PRSM & OSF & FSF & SFF & \textbf{SFF++} & PRSM & OSF & FSF & SFF & \textbf{SFF++}\Bstrut\\
		 \hline
		 Average & 15.99 & 19.84 & 15.35 & 18.15 & {\bf 13.60} & {\bf 13.70} & 28.16 & 18.32 & 22.20 & 18.47\Tstrut\Bstrut\\
		 \hline
		 alley\_1 & 7.43 & 5.28 & 5.92 & 8.81 & \textbf{3.98} & {\bf 1.58} & 7.33 & 2.11 & 3.95 & 2.11\Tstrut\\
		 alley\_2 & {\bf 0.79} & 1.31 & 2.08 & 1.73 & 1.27 & 1.08 & 1.44 & 1.20 & {\bf 0.87} & 1.01 \\
		 ambush\_2 & 41.77 & 55.13 & {36.93} & 51.72 & \textbf{31.56} & {\bf 51.33} & 87.37 & 72.68 & 83.84 & 76.00 \\
		 ambush\_4 & 24.09 & 24.05 & {23.30} & 37.78 & \textbf{22.25} & {\bf 41.99} & 49.16 & 45.23 & 42.65 & 61.88 \\
		 ambush\_5 & {17.72} & 19.54 & 18.54 & 25.52 & \textbf{13.48} & 25.23 & 44.70 & {\bf 24.82} & 29.86 & 32.96 \\
		 ambush\_6 & 29.41 & {26.18} & 30.33 & 37.13 & \textbf{23.17} & {\bf  41.98} & 54.75 & 44.05 & 47.65 & 59.26\\
		 ambush\_7 & 35.07 & 71.58 & 23.47 & {\bf 16.34} & 24.62 & {\bf 3.35} & 22.47 & 27.87 & 7.35 & 9.99\\ 
		 bamboo\_1 & {\bf 7.34} & 9.71 & 9.67 & 14.53 & 10.80 & {\bf 2.41} & 4.04 & 4.11 & 4.15 & 3.44\\
		 bamboo\_2 & {\bf 17.06} & 18.08 & 19.27 & 19.89 & 18.90 & {3.58} & 4.86 & 3.65 & 3.97 & \textbf{3.57}\\
		 bandage\_1 & 21.22 & 19.37 & 20.93 & {\bf 16.42} & 17.46 & {\bf 3.30} & 18.40 & 4.00 & 4.03 & 4.10\\
		 bandage\_2 & 22.44 & 23.53 & 22.69 & {21.77} & \textbf{16.80} & {\bf 4.06} & 13.12 & 4.76 & 9.06 & 4.56\\
		 cave\_4 & {\bf 4.27} & 5.86 & 6.22 & 6.20 & 4.93 & 16.32 & 33.94 & 14.62 & {\bf 12.95} & 18.16\\
		 market\_2 & {\bf 5.27} & 6.61 & 6.81 & 6.71 & 6.26 & {\bf 4.77} & 10.08 & 5.17 & 6.09 & 5.51\\
		 market\_5 & 15.38 & 13.67 & {\bf 13.25} & 26.66 & 14.13 & 28.38 & 29.58 & {\bf 26.31} & 28.87 & 32.56\\
		 market\_6 & {\bf 8.99} & 10.29 & 10.63 & 14.53 & 10.18 & {\bf 10.72} & 16.39 & 13.13 & 16.69 & 13.91\\
		 mountain\_1 & 0.42 & 0.78 & 0.23 & {0.15} & \textbf{0.02} & {\bf 3.71} & 88.60 & 17.05 & 89.57 & 10.84\\
		 shaman\_2 & 25.49 & 28.27 & 24.77 & {\bf 21.13} & 23.94 & {\bf 0.46} & 1.67 & 0.56 & 4.31 & 1.80\\
		 shaman\_3 & 33.92 & 52.22 & {\bf 27.09} & 35.37 & 29.02 & 1.75 & 11.45 & {\bf 1.31} & 8.51 & 5.53\\
		 sleeping\_2 & {\bf 1.74} & 2.97 & 3.52 & 3.07 & 2.24 & {\bf 0.00} & 0.01 & 0.02 & 0.03 & \textbf{0.00} \\
		 temple\_2 & {\bf 4.92} & 5.54 & 5.96 & 6.98 & 4.95 & {\bf 9.51} & 10.52 & 9.66 & 12.57 & 12.05\\
		 temple\_3 & 11.04 & 16.62 & 10.65 & 8.61 & \textbf{5.76} & 32.10 & 81.39 & 62.34 & 49.18 & \textbf{28.64}\\
		\end{tabular}
\end{table*}

\begin{figure*}[t]
	\centering
	\hspace{0.03\textwidth}%
	\begin{subfigure}[c]{0.03\textwidth}
		\rotatebox[origin=c]{90}{GT}%
	\end{subfigure}%
	\begin{subfigure}[c]{0.3\textwidth}
		\includegraphics[width=1\columnwidth]{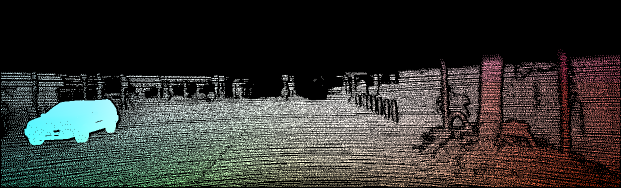}%
	\end{subfigure}
	\begin{subfigure}[c]{0.3\textwidth}
		\includegraphics[width=1\columnwidth]{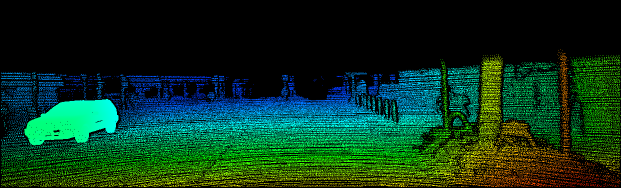}%
	\end{subfigure}
	\begin{subfigure}[c]{0.3\textwidth}
		\includegraphics[width=1\columnwidth]{img/0001}%
	\end{subfigure}\\%
	\hspace{0.03\textwidth}%
	\hspace{0.03\textwidth}%
	\begin{subfigure}[c]{0.3\textwidth}
		\centering Optical Flow
	\end{subfigure}
	\begin{subfigure}[c]{0.3\textwidth}
		\centering Disparity $t+1$
	\end{subfigure}
	\begin{subfigure}[c]{0.3\textwidth}
		\centering Image\Tstrut\\Scene Flow Error\Bstrut
	\end{subfigure}\\%
	\hspace{0.03\textwidth}%
	\begin{subfigure}[c]{0.03\textwidth}
		\rotatebox[origin=c]{90}{Matching}%
	\end{subfigure}%
	\begin{subfigure}[c]{0.3\textwidth}
		\includegraphics[width=1\textwidth]{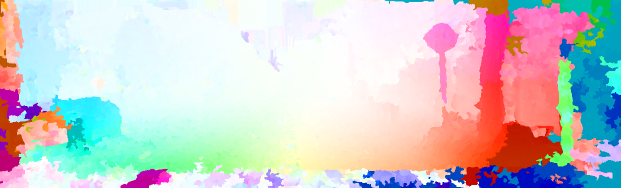}%
		\vspace{1mm}%
	\end{subfigure}
	\begin{subfigure}[c]{0.3\textwidth}
		\includegraphics[width=1\textwidth]{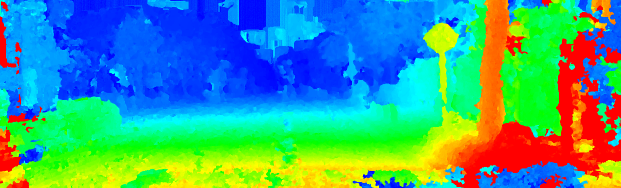}%
		\vspace{1mm}%
	\end{subfigure}
	\begin{subfigure}[c]{0.3\textwidth}
		\includegraphics[width=1\textwidth]{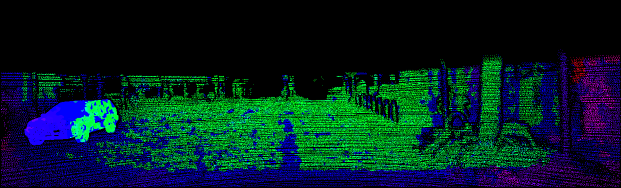}%
		\vspace{1mm}%
	\end{subfigure}\\%
	\begin{subfigure}[c]{0.03\textwidth}
		\rotatebox[origin=c]{90}{SFF}%
	\end{subfigure}%
	\begin{subfigure}[c]{0.03\textwidth}
		\rotatebox[origin=c]{90}{Filtered}%
	\end{subfigure}%
	\begin{subfigure}[c]{0.3\textwidth}
		\includegraphics[width=1\textwidth]{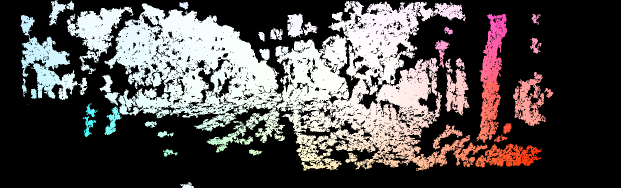}%
		\vspace{1mm}%
	\end{subfigure}
	\begin{subfigure}[c]{0.3\textwidth}
		\includegraphics[width=1\textwidth]{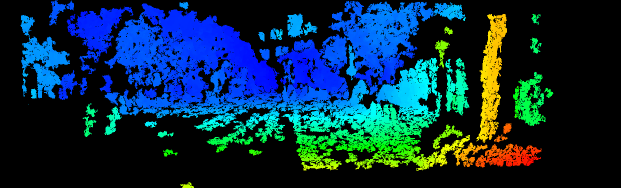}%
		\vspace{1mm}%
	\end{subfigure}
	\begin{subfigure}[c]{0.3\textwidth}
		\includegraphics[width=1\textwidth]{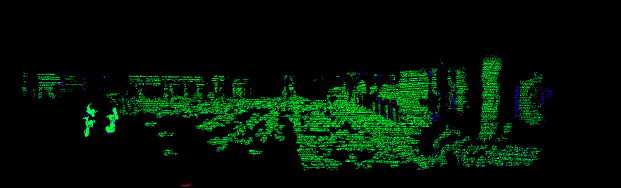}%
		\vspace{1mm}%
	\end{subfigure}\\%
	\hspace{0.03\textwidth}%
	\begin{subfigure}[c]{0.03\textwidth}
		\rotatebox[origin=c]{90}{Interp.}%
	\end{subfigure}%
	\begin{subfigure}[c]{0.3\textwidth}
		\includegraphics[width=1\textwidth]{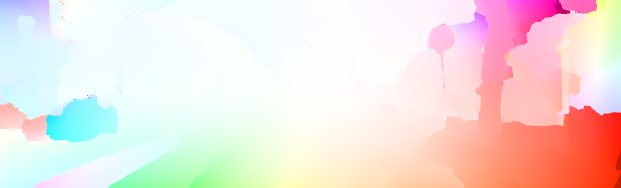}%
		\vspace{2.5mm}%
	\end{subfigure}
	\begin{subfigure}[c]{0.3\textwidth}
		\includegraphics[width=1\textwidth]{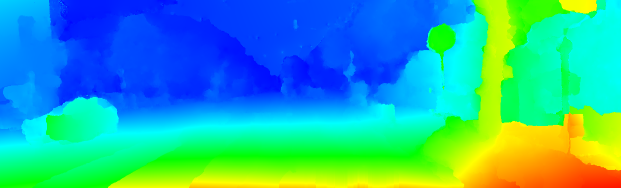}%
		\vspace{2.5mm}%
	\end{subfigure}
	\begin{subfigure}[c]{0.3\textwidth}
		\includegraphics[width=1\textwidth]{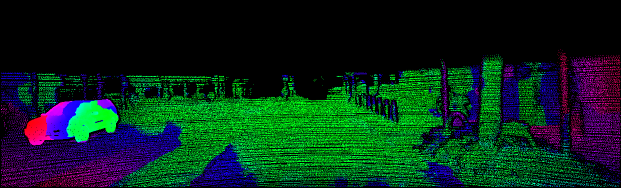}%
		\vspace{2.5mm}%
	\end{subfigure}\\%
	\hspace{0.03\textwidth}%
	\begin{subfigure}[c]{0.03\textwidth}
		\rotatebox[origin=c]{90}{Matching}%
	\end{subfigure}%
	\begin{subfigure}[c]{0.3\textwidth}
		\includegraphics[width=1\textwidth]{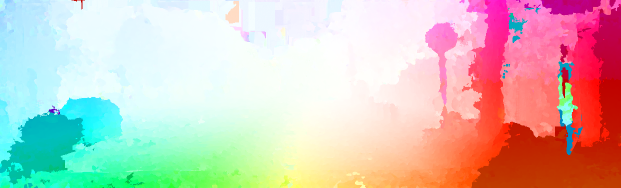}%
		\vspace{1mm}%
	\end{subfigure}
	\begin{subfigure}[c]{0.3\textwidth}
		\includegraphics[width=1\textwidth]{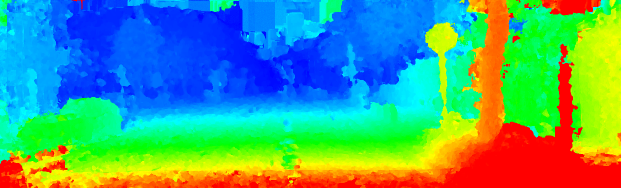}%
		\vspace{1mm}%
	\end{subfigure}
	\begin{subfigure}[c]{0.3\textwidth}
		\includegraphics[width=1\textwidth]{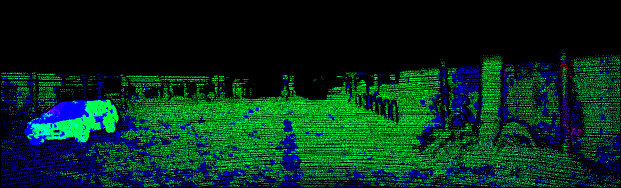}%
		\vspace{1mm}%
	\end{subfigure}\\%
	\begin{subfigure}[c]{0.03\textwidth}
		\rotatebox[origin=c]{90}{\bf \acrshort*{sffpp}}%
	\end{subfigure}%
	\begin{subfigure}[c]{0.03\textwidth}
		\rotatebox[origin=c]{90}{Filtered}%
	\end{subfigure}%
	\begin{subfigure}[c]{0.3\textwidth}
		\includegraphics[width=1\textwidth]{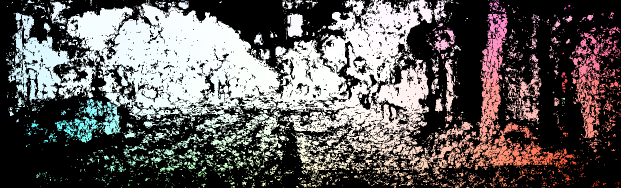}%
		\vspace{1mm}%
	\end{subfigure}
	\begin{subfigure}[c]{0.3\textwidth}
		\includegraphics[width=1\textwidth]{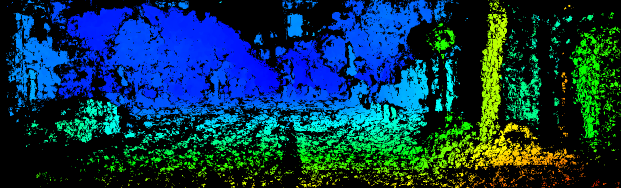}%
		\vspace{1mm}%
	\end{subfigure}
	\begin{subfigure}[c]{0.3\textwidth}
		\includegraphics[width=1\textwidth]{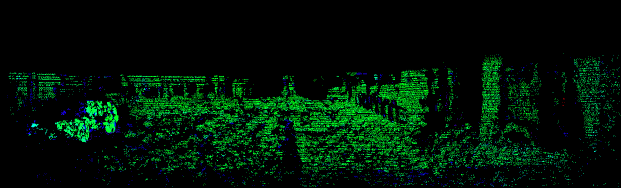}%
		\vspace{1mm}%
	\end{subfigure}\\%
	\hspace{0.03\textwidth}%
	\begin{subfigure}[c]{0.03\textwidth}
		\rotatebox[origin=c]{90}{Interp.}%
	\end{subfigure}%
	\begin{subfigure}[c]{0.3\textwidth}
		\includegraphics[width=1\textwidth]{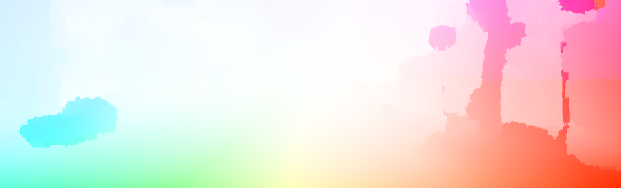}%
	\end{subfigure}
	\begin{subfigure}[c]{0.3\textwidth}
		\includegraphics[width=1\textwidth]{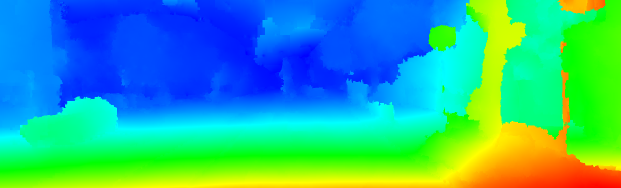}%
	\end{subfigure}
	\begin{subfigure}[c]{0.3\textwidth}
		\includegraphics[width=1\textwidth]{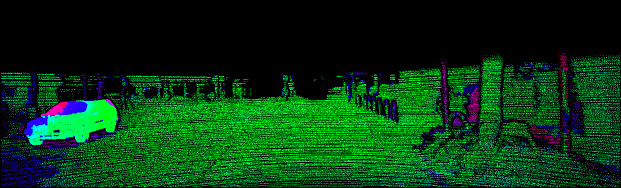}%
	\end{subfigure}
	\caption{Visual Comparison of \gls*{sff} \citep{schuster2018sceneflowfields} and our novel \gls*{sffpp}. Opical flow, disparity at $t+1$ and scene flow error maps are shown for matching, filtering, and dense interpolation.} \label{fig:dual_vs_multi}
\end{figure*}

\subsection{Ablation Study} \label{sec:results:ablation}
The modularity of our method allows us to replace or leave out various parts of the pipeline. This way, it is easy to evaluate the effect of each component separately. We do so in \cref{tab:ablation} by evaluating the results of all 200 training images of the KITTI data set \citep{menze2015object} for several variants of our method. In particular, we vary the number of frames for the matching between \textit{dual} and \textit{multi}. We also alter the interpolation mechanism from the original one presented by \citet{schuster2018sceneflowfields} (\textit{epic3d}) to the robust version presented within this work in \cref{sec:extension:interpolation} (\textit{ric3d}). Further, we evaluate the impact of the training data for the boundary detector. The original version was trained on BSDS \citep{martin2001bsds} (\textit{bsds}). \citet{schuster2018sceneflowfields} presented a specialized version optimized for the KITTI data set (\textit{kitti}). Lastly, we present results for our universal boundary detector (cf. \cref{sec:extension:interpolation}) trained on a mix of data from BSDS, KITTI, and Sintel \citep{butler2012sintel} (\textit{mixed}). Finally, we successively add the variational refinement (\textit{var}) and the ego-motion optimization (\textit{ego}) for different combinations of the components.

In conformity with previous experiments \citep{schuster2018sceneflowfields}, the dedicated KITTI boundary detector outperforms the original version that was trained on BSDS only. The universal boundary detector that was trained on mixed data performs almost equally well on KITTI and much better on other data sets like Sintel. Further, the novel more robust interpolation improves the accuracy of the scene flow result greatly, making the ego-motion model and even the variational refinement in almost all cases obsolete. Multi-frame matching improves the results even more. The impact of the multi-frame setup will be analyzed in \cref{sec:results:extension} in more detail.

\begin{figure*}[t]
	\centering
	\begin{subfigure}[c]{0.03\textwidth}
		\rotatebox[origin=c]{90}{Image}%
	\end{subfigure}
	\begin{subfigure}[c]{0.3\textwidth}
		\includegraphics[width=1\textwidth]{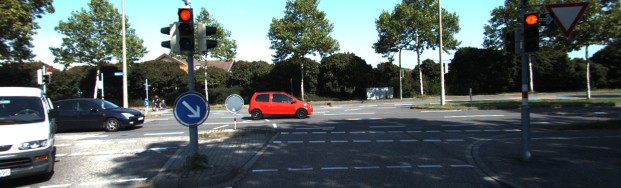}%
		\vspace{2mm}%
	\end{subfigure}
	\begin{subfigure}[c]{0.3\textwidth}
		\includegraphics[width=1\textwidth]{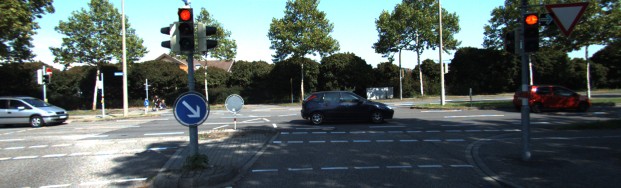}%
		\vspace{2mm}%
	\end{subfigure}
	\begin{subfigure}[c]{0.3\textwidth}
		\includegraphics[width=1\textwidth]{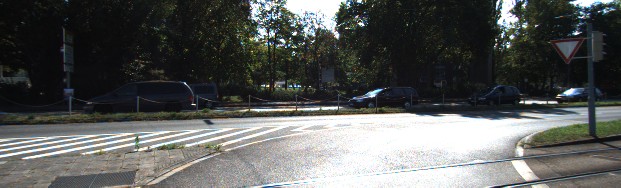}%
		\vspace{2mm}%
	\end{subfigure}\\%
	\begin{subfigure}[c]{0.03\textwidth}
		\rotatebox[origin=c]{90}{PRSM}%
	\end{subfigure}
	\begin{subfigure}[c]{0.3\textwidth}
		\includegraphics[width=1\textwidth]{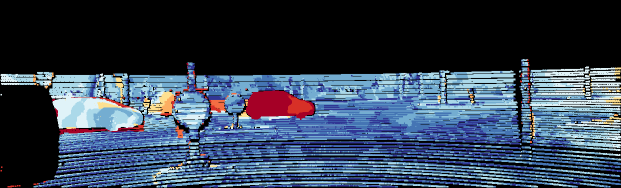}%
		\vspace{1mm}%
	\end{subfigure}
	\begin{subfigure}[c]{0.3\textwidth}
		\includegraphics[width=1\textwidth]{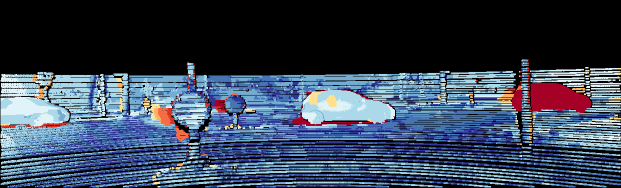}%
		\vspace{1mm}%
	\end{subfigure}
	\begin{subfigure}[c]{0.3\textwidth}
		\includegraphics[width=1\textwidth]{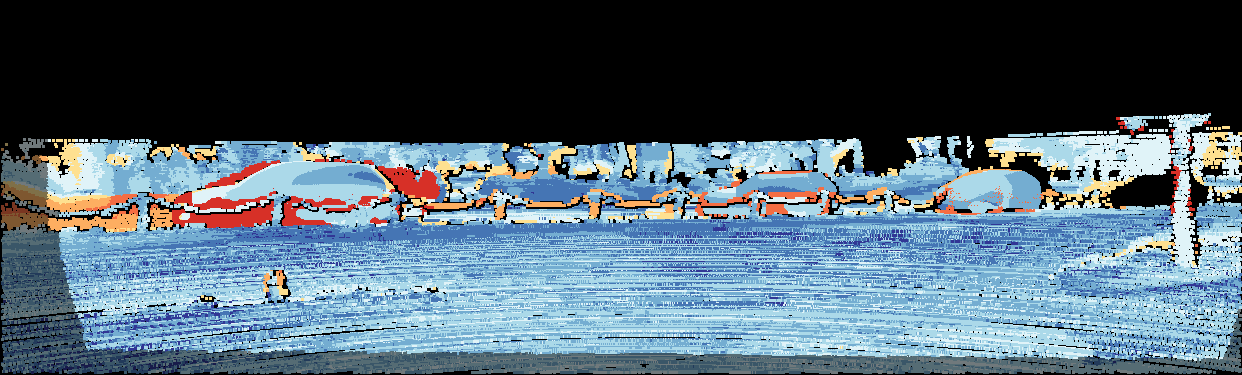}%
		\vspace{1mm}%
	\end{subfigure}\\%
	\begin{subfigure}[c]{0.03\textwidth}
		\rotatebox[origin=c]{90}{OSF+TC}%
	\end{subfigure}
	\begin{subfigure}[c]{0.3\textwidth}
		\includegraphics[width=1\textwidth]{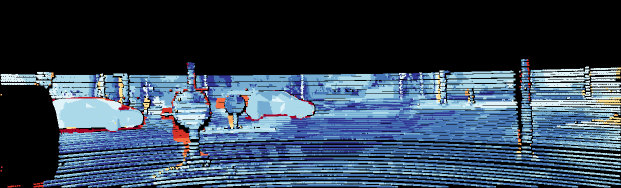}%
		\vspace{1mm}%
	\end{subfigure}
	\begin{subfigure}[c]{0.3\textwidth}
		\includegraphics[width=1\textwidth]{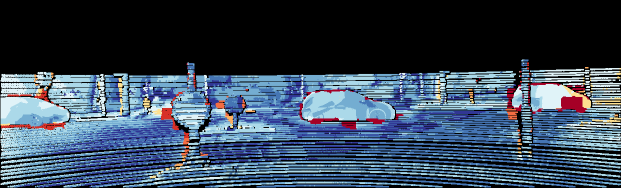}%
		\vspace{1mm}%
	\end{subfigure}
	\begin{subfigure}[c]{0.3\textwidth}
		\includegraphics[width=1\textwidth]{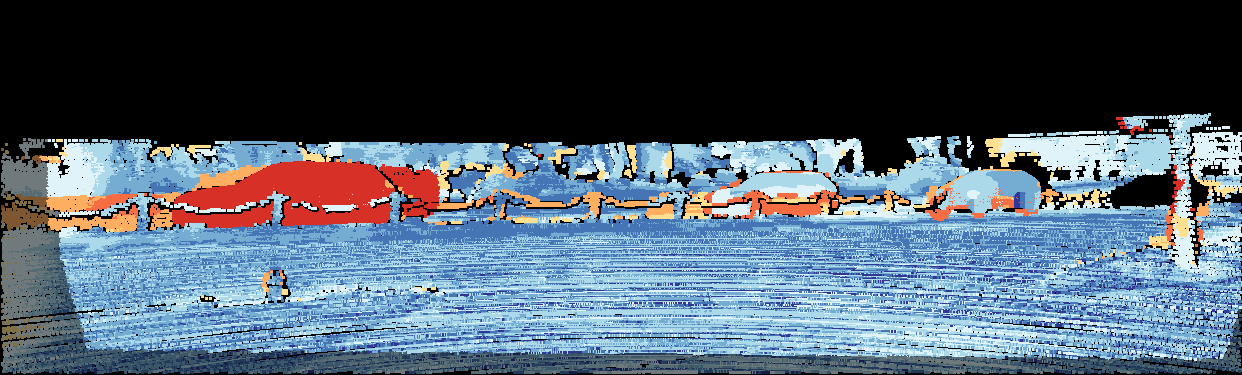}%
		\vspace{1mm}%
	\end{subfigure}\\%
	\begin{subfigure}[c]{0.03\textwidth}
		\rotatebox[origin=c]{90}{FSF+MS}%
	\end{subfigure}
	\begin{subfigure}[c]{0.3\textwidth}
		\includegraphics[width=1\textwidth]{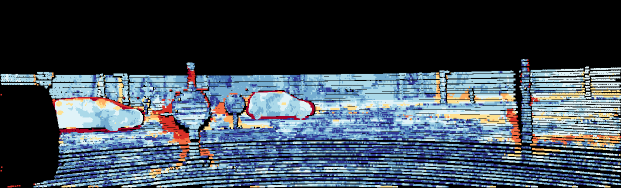}%
		\vspace{1mm}%
	\end{subfigure}
	\begin{subfigure}[c]{0.3\textwidth}
		\includegraphics[width=1\textwidth]{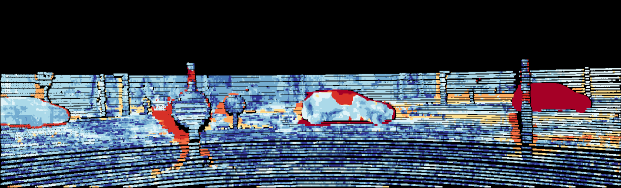}%
		\vspace{1mm}%
	\end{subfigure}
	\begin{subfigure}[c]{0.3\textwidth}
		\includegraphics[width=1\textwidth]{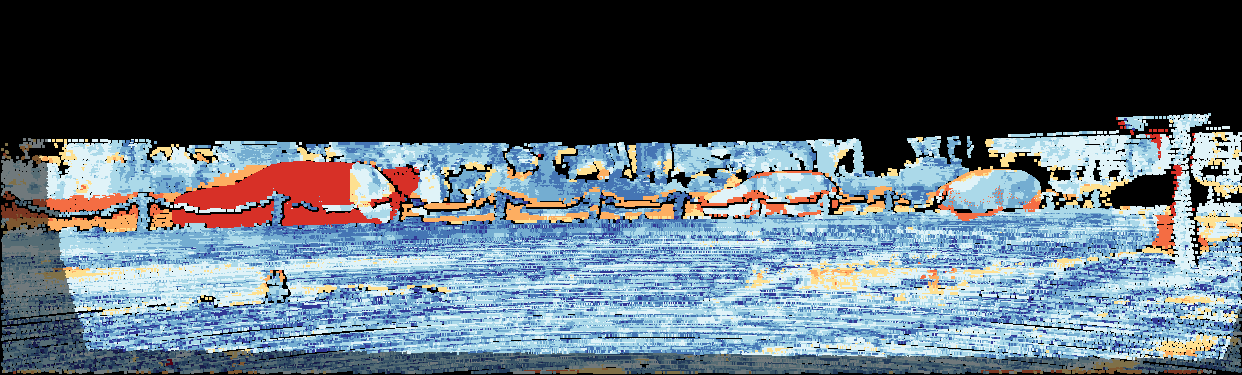}%
		\vspace{1mm}%
	\end{subfigure}\\%
	\begin{subfigure}[c]{0.03\textwidth}
		\rotatebox[origin=c]{90}{SFF}%
	\end{subfigure}
	\begin{subfigure}[c]{0.3\textwidth}
		\includegraphics[width=1\textwidth]{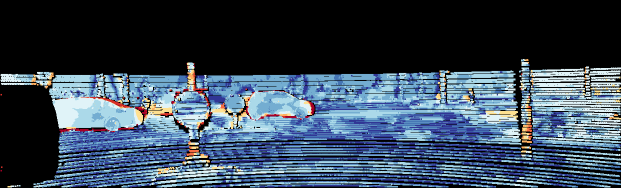}%
		\vspace{1mm}%
	\end{subfigure}
	\begin{subfigure}[c]{0.3\textwidth}
		\includegraphics[width=1\textwidth]{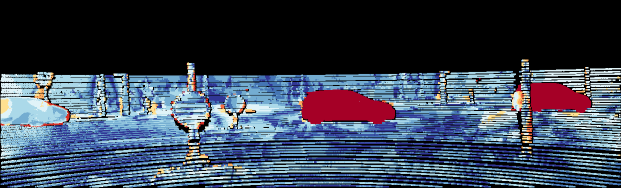}%
		\vspace{1mm}%
	\end{subfigure}
	\begin{subfigure}[c]{0.3\textwidth}
		\includegraphics[width=1\textwidth]{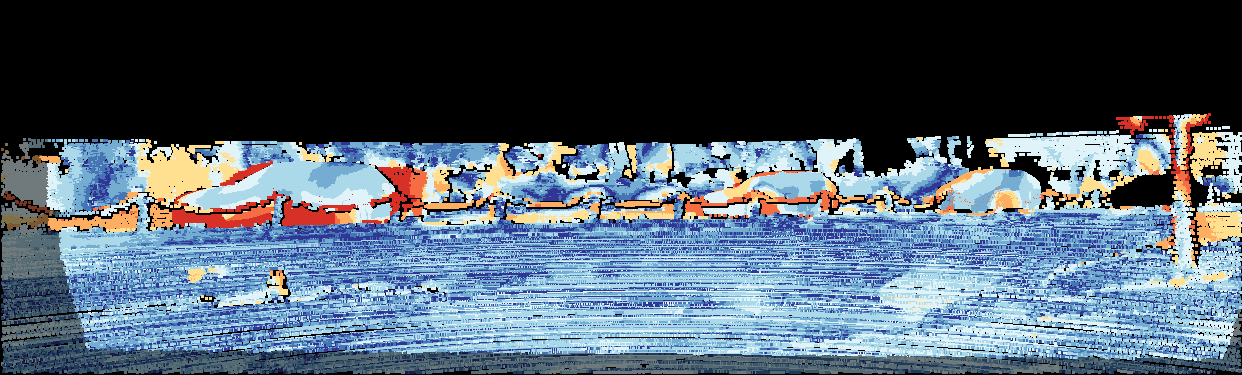}%
		\vspace{1mm}%
	\end{subfigure}\\%
	\begin{subfigure}[c]{0.03\textwidth}
		\rotatebox[origin=c]{90}{\bf SFF++}%
	\end{subfigure}
	\begin{subfigure}[c]{0.3\textwidth}
		\includegraphics[width=1\textwidth]{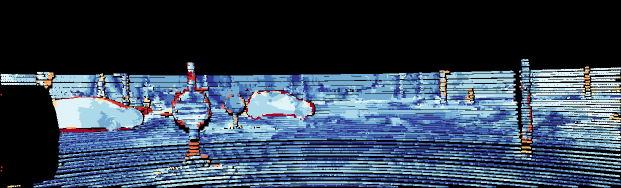}%
		\vspace{1mm}%
	\end{subfigure}
	\begin{subfigure}[c]{0.3\textwidth}
		\includegraphics[width=1\textwidth]{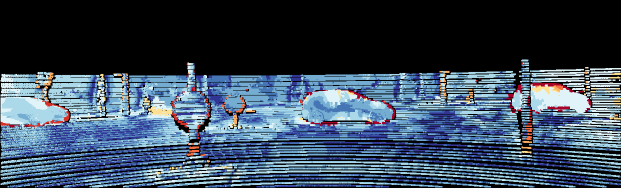}%
		\vspace{1mm}%
	\end{subfigure}
	\begin{subfigure}[c]{0.3\textwidth}
		\includegraphics[width=1\textwidth]{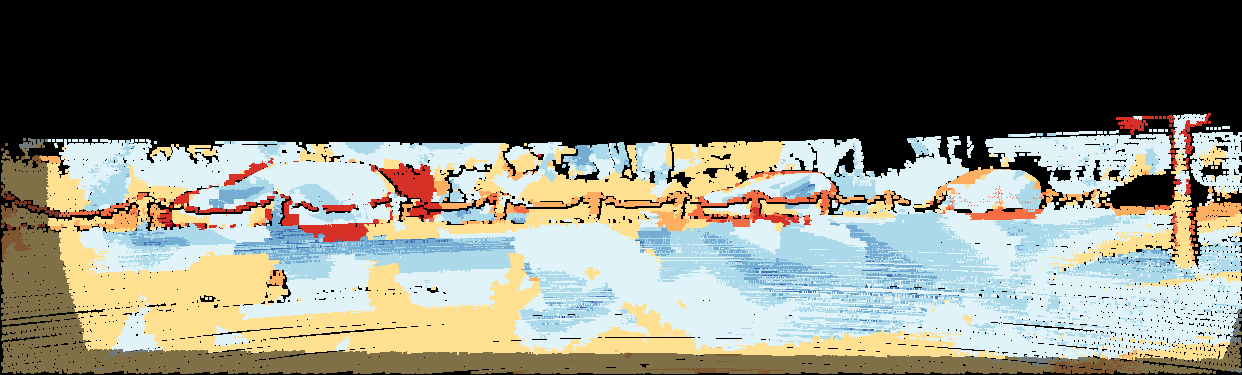}%
		\vspace{1mm}%
	\end{subfigure}\\%
	\raggedright 
	\begin{subfigure}[c]{0.1\textwidth}
		\centering
		EPE%
	\end{subfigure}
	\begin{subfigure}[c]{0.85\textwidth}
		\includegraphics[width=1\textwidth]{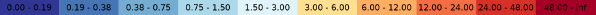}%
	\end{subfigure}
	\caption{Comparison of the scene flow error on the public KITTI Scene Flow Benchmark \citep{menze2015object} for \gls*{prsm} \citep{vogel2015PRSM}, OSF+TC \citep{menze2018osf}, FSF+MS \citep{taniai2017fsf}, \gls*{sff} \citep{schuster2018sceneflowfields}, and our novel approach \gls*{sffpp}.} 
	\label{fig:kitti}
\end{figure*}

\subsection{Evaluation of the Extensions} \label{sec:results:extension}
For visual comparison of our robust multi-frame approach \gls*{sffpp} to \gls*{sff} \citep{schuster2018sceneflowfields}, we show optical flow and disparity at the next time step along with a scene flow error map for full matching, filtered matches, and dense interpolation in \cref{fig:dual_vs_multi}. The error maps give correct estimates in green and outliers in blue to red according to the KITTI metric. Our novel multi-frame approach with explicit visibility reasoning is able to match occluded areas (e.g. next to the tree and traffic signs) and out-of-bounds regions (e.g. the front of the car) that are both not visible in some of the relevant frames (cf. \cref{fig:visibility:predict}). In addition, our robust interpolation can handle regions with almost no input seeds (e.g the lower left part of the image).
\cref{tab:dual_vs_multi} compares the same in terms of average end-point-error and average outliers on all KITTI training sequences for all ground truth pixels (KITTI \textit{occ} data) and occluded regions only (\textit{occ} without \textit{noc}). 
Our novel method is already very accurate during matching with a much smaller average end-point-error compared to \gls*{sff}. Unlike SFF, our multi-frame approach is able to match almost 40 \% of the invisible areas. Though overall filtered results after the consistency check appear to be worse, our novel approach is able to retain 9.8 \% of the matches in occluded areas where SFF filters almost everything. That makes the spatial distribution of our matches preferable to that of the dual-frame approach (cf. \cref{fig:dual_vs_multi}). Because of this and due to our robust interpolation, final results are much better with only about one third of scene flow outliers in occluded areas and less than half the percentage of overall outliers.

The direct comparison in these experiments shows that our novel multi-frame matching strategy with explicit visibility reasoning and robust interpolation is superior to the original \gls*{sff} \citep{schuster2018sceneflowfields}. Even with the significant boost of the optional ego-motion model of \citet{schuster2018sceneflowfields}, our novel method without this model outperforms the dual-frame version on both evaluated data sets (see \cref{sec:results:kitti,sec:results:sintel}).

\subsection{KITTI Benchmark} \label{sec:results:kitti}
We use the famous KITTI data set \citep{menze2015object} for the first part of our comparison to state-of-the-art. Please note that this benchmark does not provide reference motions for vulnerable road users like pedestrians or cyclists. The only dynamic objects are rigidly moving vehicles.
Complete results for this benchmark are publicly available on the respective homepage. In \cref{tab:kitti} we give the top methods of the leader board, where the average amount of outliers are compared for disparity at both time steps (\textit{D1}, \textit{D2}), optical flow (\textit{Fl}), and scene flow (\textit{SF}). Each category is further divided into regions of background (\textit{bg}, static areas), foreground (\textit{fg}, moving objects), and both (\textit{all}, all available ground truth).
As assumed, our novel approach scores better than the dual-frame conference version \citep{schuster2018sceneflowfields}. Especially for the important foreground regions, our results are even comparable to the deep learning approach of \citet{ren2017cascaded} that uses semantic segmentation and to that of \citet{menze2015object} that explicitly estimates a single rigid motion for complete independent objects. In addition, our method is at least three times faster than any better performing algorithm.

While investigating the discrepancy between our results on the validation data (\cref{tab:dual_vs_multi}) and on the test data (\cref{tab:kitti}), we discovered that our multi-frame approach has difficulties in case the constant motion assumption is harshly violated (cf. \cref{sec:results:limitations}). This can happen in KITTI due to the low frame rate of 10 frames per second and because of strong sudden pitch or roll rotations on bumpy streets, \eg the last example in \cref{fig:kitti}.
However, in most cases the assumption holds and produces robust and accurate results.

We compare the error maps for different frames of the KITTI test data for \gls*{prsm}, OSF+TC, FSF+MS, SFF, and our novel method SFF++ in \cref{fig:kitti}. Even for the partly occluded, badly illuminated vehicle in the second example, scene flow is estimated reliably. In this example also, scene flow at the occluded areas around the left traffic light is predicted correctly with sharp boundaries around the traffic sign. Within the third example, larger parts of the background exceed the error threshold slightly due to the violated constant motion assumption. However, dynamic objects are still detected more reliably than in most other approaches.

\begin{figure}[t]
	\centering	
	\begin{subfigure}[c]{0.04\columnwidth}
		\rotatebox[origin=c]{90}{Image}
	\end{subfigure}
	\begin{subfigure}[c]{0.4\columnwidth}
		\includegraphics[width=1\columnwidth]{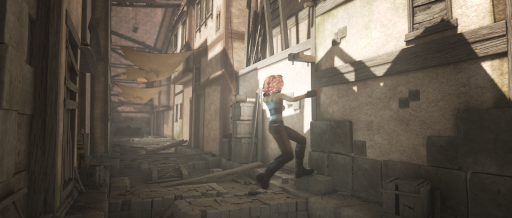}%
		\vspace{1mm}%
	\end{subfigure}
	\begin{subfigure}[c]{0.4\columnwidth}
		\includegraphics[width=1\columnwidth]{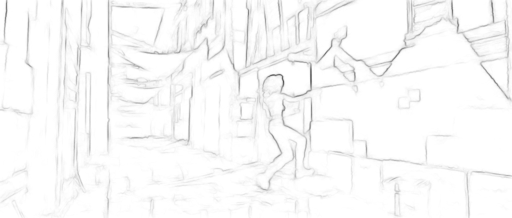}%
		\vspace{1mm}%
	\end{subfigure}
	\begin{subfigure}[c]{0.04\columnwidth}
		\rotatebox[origin=c]{90}{Edges}
	\end{subfigure}\\%
	\begin{subfigure}[c]{0.4\columnwidth}
		\centering Flow
	\end{subfigure}
	\begin{subfigure}[c]{0.4\columnwidth}
		\centering Disparity
	\end{subfigure}\\%
	\begin{subfigure}[c]{0.04\columnwidth}
		\rotatebox[origin=c]{90}{Estimate}
	\end{subfigure}
	\begin{subfigure}[c]{0.4\columnwidth}
		\includegraphics[width=1\columnwidth]{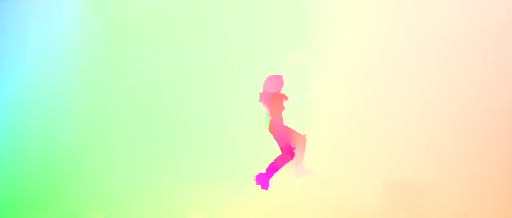}%
		\vspace{0.5mm}%
	\end{subfigure}
	\begin{subfigure}[c]{0.4\columnwidth}
		\includegraphics[width=1\columnwidth]{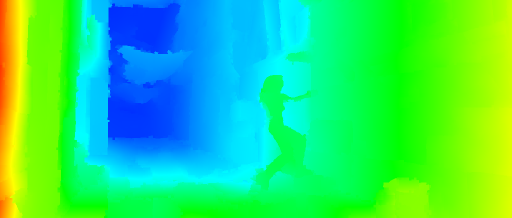}%
		\vspace{0.5mm}%
	\end{subfigure}
	\begin{subfigure}[c]{0.04\columnwidth}
		\rotatebox[origin=c]{90}{Estimate}
	\end{subfigure}\\%
	\begin{subfigure}[c]{0.04\columnwidth}
		\rotatebox[origin=c]{90}{Error}
	\end{subfigure}
	\begin{subfigure}[c]{0.4\columnwidth}
		\includegraphics[width=1\columnwidth]{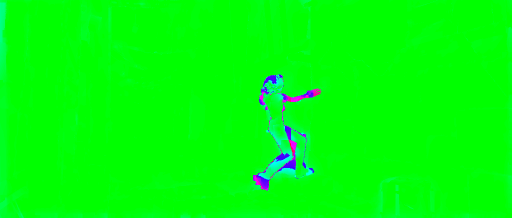}%
		\vspace{0.5mm}%
	\end{subfigure}
	\begin{subfigure}[c]{0.4\columnwidth}
		\includegraphics[width=1\columnwidth]{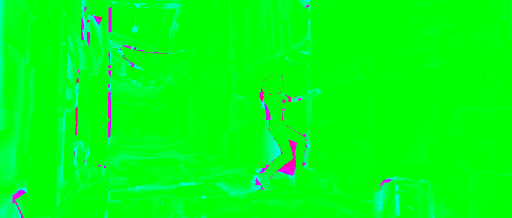}%
		\vspace{0.5mm}%
	\end{subfigure}
	\begin{subfigure}[c]{0.04\columnwidth}
		\rotatebox[origin=c]{90}{Error}
	\end{subfigure}\\%
	\begin{subfigure}[c]{0.04\columnwidth}
		\rotatebox[origin=c]{90}{GT}
	\end{subfigure}
	\begin{subfigure}[c]{0.4\columnwidth}
		\includegraphics[width=1\columnwidth]{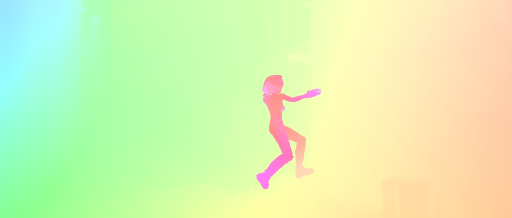}%
		\vspace{3mm}%
	\end{subfigure}
	\begin{subfigure}[c]{0.4\columnwidth}
		\includegraphics[width=1\columnwidth]{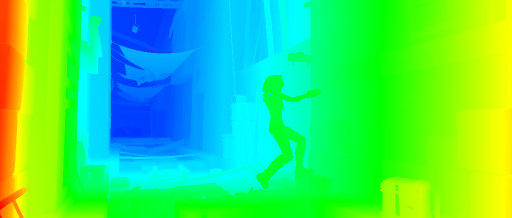}%
		\vspace{3mm}%
	\end{subfigure}
	\begin{subfigure}[c]{0.04\columnwidth}
		\rotatebox[origin=c]{90}{GT}
	\end{subfigure}\\%
	\begin{subfigure}[c]{0.04\columnwidth}
		\rotatebox[origin=c]{90}{Image}
	\end{subfigure}
	\begin{subfigure}[c]{0.4\columnwidth}
		\includegraphics[width=1\columnwidth]{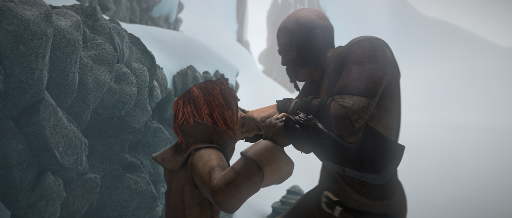}%
		\vspace{1mm}%
	\end{subfigure}
	\begin{subfigure}[c]{0.4\columnwidth}
		\includegraphics[width=1\columnwidth]{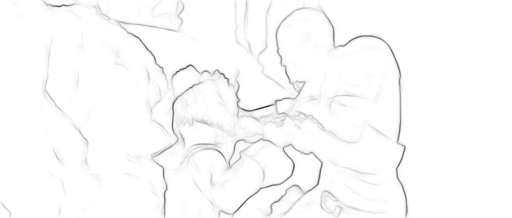}%
		\vspace{1mm}%
	\end{subfigure}
	\begin{subfigure}[c]{0.04\columnwidth}
		\rotatebox[origin=c]{90}{Edges}
	\end{subfigure}\\%
	\begin{subfigure}[c]{0.4\columnwidth}
		\centering Flow
	\end{subfigure}
	\begin{subfigure}[c]{0.4\columnwidth}
		\centering Disparity
	\end{subfigure}\\%
	\begin{subfigure}[c]{0.04\columnwidth}
		\rotatebox[origin=c]{90}{Estimate}
	\end{subfigure}
	\begin{subfigure}[c]{0.4\columnwidth}
		\includegraphics[width=1\columnwidth]{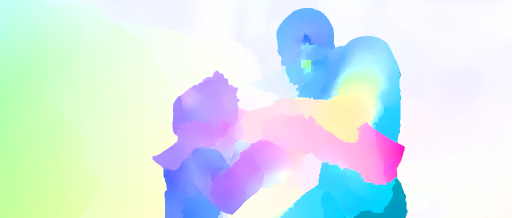}%
		\vspace{0.5mm}%
	\end{subfigure}
	\begin{subfigure}[c]{0.4\columnwidth}
		\includegraphics[width=1\columnwidth]{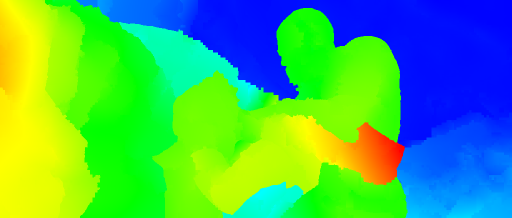}%
		\vspace{0.5mm}%
	\end{subfigure}
	\begin{subfigure}[c]{0.04\columnwidth}
		\rotatebox[origin=c]{90}{Estimate}
	\end{subfigure}\\%
	\begin{subfigure}[c]{0.04\columnwidth}
		\rotatebox[origin=c]{90}{Error}
	\end{subfigure}
	\begin{subfigure}[c]{0.4\columnwidth}
		\includegraphics[width=1\columnwidth]{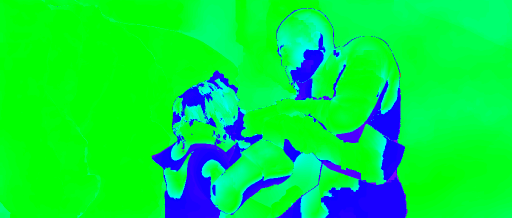}%
		\vspace{0.5mm}%
	\end{subfigure}
	\begin{subfigure}[c]{0.4\columnwidth}
		\includegraphics[width=1\columnwidth]{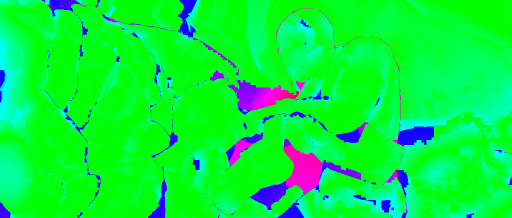}%
		\vspace{0.5mm}%
	\end{subfigure}
	\begin{subfigure}[c]{0.04\columnwidth}
		\rotatebox[origin=c]{90}{Error}
	\end{subfigure}\\%
	\begin{subfigure}[c]{0.04\columnwidth}
		\rotatebox[origin=c]{90}{GT}
	\end{subfigure}
	\begin{subfigure}[c]{0.4\columnwidth}
		\includegraphics[width=1\columnwidth]{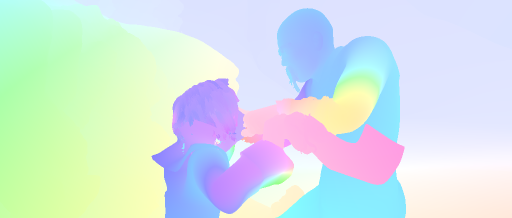}%
	\end{subfigure}
	\begin{subfigure}[c]{0.4\columnwidth}
		\includegraphics[width=1\columnwidth]{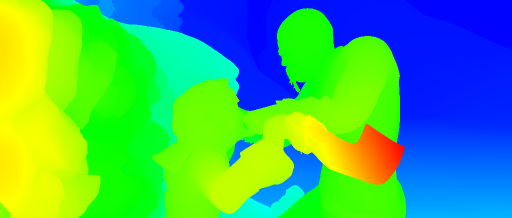}%
	\end{subfigure}
	\begin{subfigure}[c]{0.04\columnwidth}
		\rotatebox[origin=c]{90}{GT}
	\end{subfigure}
	\caption{Exemplary results of our method on the Sintel data set \citep{butler2012sintel}.}
	\label{fig:sintel}
\end{figure}

\subsection{MPI Sintel} \label{sec:results:sintel}
The impact of our contributions has already been shown on a public benchmark in the previous section.
To additionally demonstrate the robustness and versatility, we evaluate \gls*{sff} and our more robust multi-frame extension \gls*{sffpp} on a second data set -- MPI Sintel \citep{butler2012sintel}. For our novel \gls*{sffpp}, we use exactly the same set of parameters. We explicitly highlight this, since many methods rely on parameter tuning. Especially deep learning models often require fine tuning.
Sintel has a lot of contrary properties to KITTI. Most prominently, MPI Sintel consists of non-realistic, synthetically rendered images. Further, images are captured from a totally different domain and have therefore different characteristics. Sintel contains small as well as very large motion displacements of deformable, articulated characters. Thus, many of the included motions are non-rigid and the geometries are less often planar, compared to KITTI. This is also reflected in the results of this data set that are given in \cref{tab:sintel}. OSF \citep{menze2015object}, that is strongly relying on the piece-wise rigid plane model, performs considerably worse than on KITTI. Same is assumed for methods that rely on specialized deep neural networks, e.g. \citep{behl2017bounding,ren2017cascaded}.

For evaluation, we have used the \textit{final} rendering pass of all training sequences except \textit{cave\_2} and \textit{sleeping\_1} as \citet{taniai2017fsf,schuster2018sceneflowfields}. Since no full scene flow ground truth is available, we compute the average percentage of outliers for disparity and optical flow, similar to \textit{D1-all} and \textit{Fl-all} on KITTI.
\Gls*{sffpp} outperforms the less robust dual-frame version \gls*{sff} \citep{schuster2018sceneflowfields}, again highlighting the improvements of our contribution. We even achieve the best accuracy for disparity with quite some margin to the next best method FSF \citep{taniai2017fsf} and the best performing method on KITTI, \gls*{prsm} \citep{vogel2015PRSM}. 
With that, our method joins the few methods with top performance on both data sets next to \gls*{prsm} \citep{vogel2015PRSM}, FSF+MS \citep{taniai2017fsf}, and SFF \citep{schuster2018sceneflowfields}.

\cref{fig:sintel} shows exemplary results of our approach for one frame of the sequences \textit{alley\_2} and \textit{ambush\_5}. The latter can be considered particularly challenging (cf. \cref{tab:sintel}), though it is handled quite well by our approach. \cref{fig:sintel} also demonstrates that a bit of accuracy is traded in for increased robustness. Small details, \eg the arm of the girl, are lost during interpolation.

\subsection{Limitations} \label{sec:results:limitations}
Despite all efforts, it is not possible to avoid any assumption to estimate scene flow due to the ill-posedness of the problem. However, we tried to keep the method as unrestricted as possible.
Our initial matching is purely data-based and requires no assumptions at all.
The later imposed regularization during interpolation makes use of the piece-wise rigid plane model but applies it to an extreme over-segmentation of the scene into tiny superpixels of 25 pixels to reduce the negative effects greatly.
Further, to make use of additional image information from multiple time steps, \gls{sffpp} assumes a constant motion within a temporal window of three stereo frames.
With that, \gls{sffpp} is subject to a set of (mostly theoretical) limitations that will be described here.

As discussed several times, \gls{sffpp} keeps unaffected by the drawbacks of a piece-wise rigid, planar motion assumption even though this model is used during interpolation. The reason is the size of the superpixel segments. An impression of the size of the superpixels is give in \cref{fig:interpolation}. During all our experiments, we could not observe that our superpixels are the limiting factor in the estimation of non-planar surfaces or non-rigid motions. However in theory, not all kinds of objects and motions can be approximated by small superpixels.

Another hypothetical failure case arises from the complete separation of matching and regularization. If unregularized matching leads to the removal of entire image regions during the consistency check, the later imposed regularization of the interpolation can not recover the content of these regions. This phenomenon was regularly observed for the dual-frame approach when highly dynamic objects where leaving the image domain. At the same time, this case was one of the motivations for the shift to multiple frames. In the multi-frame scenario, it is much less likely that matching fails consistently for entire regions. This is also supported by the study of visible areas in \cref{fig:visibility:gt}.

Lastly, the constant motion assumption is the only assumption-based limitation that causes practical impact on the performance of \gls{sffpp}.
For the KITTI data, we could notice degradation of the estimated scene flow due to the violation of this assumption. One of these examples is shown in the right column of \cref{fig:kitti}. Anyway, this problem was encountered rarely, mostly in the presence of potholes or crossing rails that lead to an extreme sudden discontinuity of the motion.

\section{Conclusion} \label{sec:conclusion}
\gls*{sff} is the first approach that uses sparse-to-dense interpolation for scene flow estimation. The extension to multi-frame matching with visibility prediction of \gls*{sffpp} produces accurate correspondences even for occluded parts of the scene. A consistency check removes outliers. Robustly estimated models for 3D geometry and motion on small superpixels are used as novel interpolation method to reconstruct a dense scene flow field.
Results on two diverse data sets that were computed with the exact same parameters have demonstrated the robustness for both, \gls*{sff} and \gls*{sffpp}. Our robust multi-frame approach achieves the seventh best position in the overall ranking on KITTI and is among the top performing methods on MPI Sintel, with the lowest amount of outliers for disparity estimation. In a joint score on both data sets, we claim to achieve the second best performance after \gls*{prsm} \citep{vogel2015PRSM}. Separate fine tuning on each data set is assumed to further increase the accuracy of our method. However, the focus in this work was on robustness across different domains.
To make our approach even more robust in the future, we suggest to use more than three frame pairs and most importantly evolve from our constant motion assumption to a higher order motion model.


\bibliographystyle{spbasic}      
\bibliography{bib}   


\end{document}